\documentclass[10pt,journal,compsoc]{IEEEtran}

\usepackage{graphicx}

\DeclareGraphicsExtensions{.pdf,.png,.jpg}

\usepackage{subfigure}
\usepackage[subfigure]{graphfig}
\usepackage{array}
\usepackage{amsmath}
\usepackage{multirow}
\graphicspath{ {./figs/} }

\begin{document}

\title
{
    Image Restoration Using  Convolutional Auto-encoders with Symmetric Skip Connections
}

\author
{
	Xiao-Jiao Mao, Chunhua Shen, Yu-Bin Yang
	\IEEEcompsocitemizethanks
	{
		\IEEEcompsocthanksitem
        X.-J. Mao and Y.-B. Yang are with the State Key Laboratory
		for Novel Software Technology, Nanjing University, China.
        E-mail: {\tt xjmgl.nju@gmail.com, yangyubin@nju.edu.cn}.
		\IEEEcompsocthanksitem
        C. Shen is with the School of Computer Science,
		University of Adelaide, Australia.
        E-mail: {\tt chunhua.shen@adelaide.edu.au}.
        \IEEEcompsocthanksitem
        X.-J. Mao's contribution was made when visiting University of Adelaide.
        Correspondence should be addressed to C. Shen.
	}
}

\markboth{Manuscript}%
{Mao \MakeLowercase{\textit{et al.}}: Image Restoration Using  Convolutional Auto-encoders}

\IEEEtitleabstractindextext{
\begin{abstract}

    Image restoration, including image denoising, super resolution,  inpainting,
    and so on, is a well-studied problem in computer vision and image processing,
as well as a test bed for low-level image modeling algorithms. In this work, we propose
a very deep fully convolutional  auto-encoder  network  for image
restoration, which is a
 encoding-decoding framework with symmetric con\-vo\-lu\-tion\-al-de\-con\-vo\-lutional layers.
 In other words, the network is composed of multiple layers of convolution and de-convolution
operators, learning end-to-end mappings from corrupted images to the original ones. The
convolutional layers  capture the abstraction of image
contents while eliminating corruptions. Deconvolutional layers have the capability to upsample the
feature maps and  recover
the image details. To deal with the problem that deeper networks tend to be more difficult
to train, we propose to symmetrically link convolutional and deconvolutional layers with
skip-layer connections, with which the training converges much faster and attains better results.
The skip connections from convolutional layers to their mirrored corresponding
deconvolutional layers exhibit two main advantages. First, they allow the signal to be back-propagated
to bottom layers directly, and thus tackles the problem of gradient vanishing, making training deep
networks easier and achieving restoration performance gains consequently. Second, these skip
connections pass image details from convolutional layers to deconvolutional layers, which is
beneficial in recovering the clean image. Significantly, with the large capacity, we show it is
possible to cope with
different levels of corruptions using a single model.
Using the same framework, we train models on tasks of image denoising,  super resolution
removing JPEG compression artifacts, non-blind image deblurring and image inpainting.
Our experiment results on  benchmark
datasets show that our network can achieve best reported performance on all of the four tasks, and
set new state-of-the-art.

\end{abstract}

}

\maketitle
\IEEEdisplaynontitleabstractindextext
\IEEEpeerreviewmaketitle

\tableofcontents
\clearpage

\section{Introduction}

Image restoration~\cite{DBLP:conf/iccv/MairalBPSZ09,DBLP:journals/tip/DongZSL13,
DBLP:journals/pami/SchmidtJNRR16,DBLP:conf/cvpr/SchmidtR14,DBLP:conf/iccv/ZoranW11}
is a classical problem in low-level vision, which has been widely studies in the literature.
Yet, it remains an active research topic and provides a test bed for many image modeling techniques.

Generally speaking, image restoration is the operation of taking a corrupted image and
estimating the original image, which is known to be an ill-posed inverse problem. A corrupted
image $Y$ can be represented as
\begin{equation}
    y = H(x) + n
\end{equation}
where $x$ is the clean version of $y$;  $H$ is the degradation function and $n$ is the additive
noise. By accommodating different types of degradation operators and noise distributions, the same mathematical model
applies to most  low-level imaging problems such as image denoising~\cite{DBLP:conf/cvpr/LiuXZG15,
DBLP:conf/iccv/ChenZY15,DBLP:conf/iccv/XuZZZF15,DBLP:conf/cvpr/GuZZF14}, super-resolution
    \cite{DBLP:conf/accv/TimofteSG14,DBLP:conf/cvpr/YangLC13,DBLP:conf/cvpr/ZhuZY14,DBLP:conf/cvpr/ZhuZ0Y15,
DBLP:conf/iccv/RieglerSRB15,DBLP:conf/iccv/GuZXMFZ15,DBLP:conf/iccv/WangLYHH15}, inpainting
    \cite{DBLP:conf/nips/XieXC12,DBLP:journals/ijcv/RothB09,DBLP:journals/tip/MairalES08} and recovering raw
images from compressed images~\cite{DBLP:conf/iccv/DongDLT15,DBLP:journals/tip/FoiKE07,DBLP:conf/eccv/JancsaryNR12}.
In the past decades, extensive studies have been carried out to develop various of image restoration methods.

    Recently, deep neural networks (DNNs) have shown their superior performance in image processing
and computer vision tasks, ranging from high-level recognition, semantic segmentation to low-level
denoising and super-resolution. One of the early deep learning models which has been used for image
denoising is the Stacked Denoising Auto-encoders (SdA)~\cite{DBLP:conf/icml/VincentLBM08}. It is
an extension of the stacked auto-encoder~\cite{DBLP:conf/nips/BengioLPL06} and was originally designed
for unsupervised feature learning.  Denoising auto-encoders can be stacked to form a deep network
by feeding output of the previous layer to the current layer as input. Jain and Seung~\cite{DBLP:conf/nips/JainS08}
proposed to use Convolutional Neural Networks (CNN) to denoise natural images. Their framework is
the same as the recent Fully Convolutional Neural Networks (FCN) for semantic image segmentation \cite{DBLP:conf/cvpr/LongSD15}
and other tasks such as super-resolution \cite{DBLP:journals/pami/DongLHT16}, although their network
is not as deep as today's models. Their network accepts an image as the input and produces an
entire image as the output through four hidden layers of convolutional filters.
The weights are learned by
minimizing the difference between the output and the clean image.

By observing recent superior performance of CNN on image processing tasks, here we propose a very deep fully convolutional
CNN-based framework for image restoration. The input of our framework is a corrupted image, and
the output is its clean version.  We observe that it is beneficial to train a very deep model for
low-level tasks like denoising, super-resolution and JPEG deblocking. Our network is much
deeper compared to that in \cite{DBLP:conf/nips/JainS08} and recent low-level image processing models
such as \cite{DBLP:journals/pami/DongLHT16}.  Instead of using image priors, the proposed framework
learns  fully convolutional and deconvolutional mappings from corrupted images to the clean
ones in an end-to-end fashion. The network is composed of multiple layers of convolution and deconvolution operators.
As deeper networks tend to be more difficult to train, we further propose to symmetrically link convolutional
and deconvolutional layers with multiple skip-layer connections, with which the training converges much faster
and better performance is achieved.

Our main contributions can be summarized as follows.

\begin{itemize}
  
\item We propose a very deep network architecture for image restoration. The network consists of a
chain of symmetric convolutional layers and deconvolutional layers. The convolutional layers act as
the feature extractor which encode the primary components of image contents while eliminating the
corruptions. The deconvolutional layers then decode the image abstraction to recover the image content
details. To the best of our knowledge, the proposed framework is the first attempt to used both
convolution and deconvolution for low-level image restoration.

\item To better train the deep network, we propose to add skip connections between corresponding
convolutional and deconvolutional layers. These shortcuts divide the network into several blocks.
These skip connections help to back-propagate the gradients to bottom layers and pass image details
to the top layers. These two characteristics make training of the end-to-end mapping from corrupted image
to the clean one easier and more effective, and thus achieve performance improvement while the network going deeper.

\item We apply the same network for tasks such as image denoising, image super-resolution, JPEG
    deblocking, non-blind image deblurring and image inpainting.
Experiments on a few widely-used  benchmark datasets demonstrate the advantages of our network over
other recent state-of-the-art methods. Moreover, relying on the large capacity and fitting
ability, our network can be trained to obtain good restoration performance on different levels
of corruption even using a single model.
\end{itemize}

The remaining content
is organized as follows. We provide a brief review of related work in Section \ref{sec:related}.
We present the architecture of the proposed network, as well as training, testing details in Section \ref{sec:main}.
In Section \ref{sec:disc}, we discuss some relevant issues.
Experimental results and analysis are provided in Section \ref{sec:exp}.

\section{Related work}
\label{sec:related}

In the past decades, extensive studies have been conducted to develop various
 image restoration methods. See detailed reviews in a survey~\cite{DBLP:journals/spm/Milanfar13}.
Traditional methods such as the BM3D algorithm~\cite{DBLP:journals/tip/DabovFKE07}
and dictionary learning based methods~\cite{DBLP:journals/tip/WangYWCYH15,
DBLP:conf/iccv/GuZXMFZ15,DBLP:journals/tip/ChatterjeeM09} have shown very promising
performance on image restoration topics such as image denoising and super-resolution.
Due to the ill-posed nature of image restoration, image prior knowledge formulated
as regularization techniques are widely used. Classic regularization models, such as
total variation~\cite{Rudin:1992:NTV:142273.142312,Chan05TV,Oli09TV}, are effective
in removing noise artifacts, but also tend to over-smooth the images. As an alternative,
sparse representation~\cite{DBLP:journals/tip/EladA06,DBLP:journals/tip/DongZSL13,
DBLP:journals/tip/DongZSW11,DBLP:journals/pami/KimK10,DBLP:journals/tip/YangWHM10} based
prior modeling is   popular, too. Mathematically, the sparse representation model
assumes that a data point $x$ can be linearly reconstructed by an over-completed dictionary,
and most of the coefficients are zero or close to zero.

An active (and probably more promising) category for image restoration is the neural
network based methods. The most significant difference between neural network methods
and other methods is that they typically learn parameters for image restoration directly
from training data (e.g., pairs of clean and corrupted images) rather than relying on
pre-defined image priors.

Stacked denoising auto-encoder~\cite{DBLP:conf/icml/VincentLBM08} is one of the most well-known
deep neural network models which can be used for image denoising. Unsupervised pre-training,
which minimizes the reconstruction error with respect to inputs, is done for one layer at a time.
Once all layers are pre-trained, the network goes through a fine-tuning stage. Xie et al.~\cite{DBLP:conf/nips/XieXC12}
combined sparse coding and deep networks pre-trained with denoising auto-encoder for
low-level vision tasks such as image denoising and inpainting. The main idea is that the
sparsity-inducing term for regularization is proposed for improved performance. Deep network
cascade (DNC)~\cite{DBLP:conf/eccv/CuiCSZC14} is a cascade of multiple stacked collaborative
local auto-encoders for image super-resolution.  High frequency texture enhanced image patches are fed
into the network to suppress the noises and collaborate the compatibility of the overlapping patches.

Other neural network based image restoration methods using networks such as multi-layer perceptron.
Early works, such as a multi-layer perceptron with a multilevel sigmoidal function
~\cite{DBLP:journals/tsp/SivakumarD93}, have been proposed and proved to be effective in image
restoration tasks. Burger et al.~\cite{DBLP:conf/cvpr/BurgerSH12} presented a patch-based algorithm
learned on a large dataset with a plain multi-layer perceptron and is able to compete with the
state-of-the-art traditional image denoising methods such as BM3D. They also concluded that with large networks,
large training data, neural networks can achieve state-of-the-art image denoising performance,
which is confirmed in the work here.

Compared to auto-encodes and multilayer perceptron, it seems that convolutional neural networks
have achieved even more significant success in the field of image restoration. Jain and Seung
~\cite{DBLP:conf/nips/JainS08} proposed fully convolutional CNN for denoising. The network is
trained by minimizing the loss between a clean image and its corrupted version by adding noises
on it. They found that CNN works well on both blind and non-blind image denoising, providing
comparable or even superior performance to wavelet and Markov Random Field (MRF) methods.
Recently, Dong et al.~\cite{DBLP:journals/pami/DongLHT16} proposed to directly learn an end-to-end
mapping between the low/high-resolution images for image super-resolution. They observed that
convolutional neural networks are enseentially related to sparse coding based methods, i.e.,
the three layers in their network can be viewed as patch representation extractor, non-linear mapping
and image reconstructor. They also proposed variant networks for other image restoration
tasks such as JPEG debloking~\cite{DBLP:conf/iccv/DongDLT15}. Wang et al.~\cite{DBLP:conf/iccv/WangLYHH15}
argued that domain expertise represented by the conventional sparse coding is still valuable
and can be combined to achieve further improved results in image super-resolution. Instead of
training with different levels of scaling factors, they proposed to use a cascade structure to
repeatedly enlarge the low-resolution image by a fixed scale until reaching a desired size.
In general, DNN-based methods learn restoration parameters directly from data, which tends to
been more effective in real-world image restoration applications.

\begin{figure*}
\centering
\includegraphics[width=0.85\textwidth]{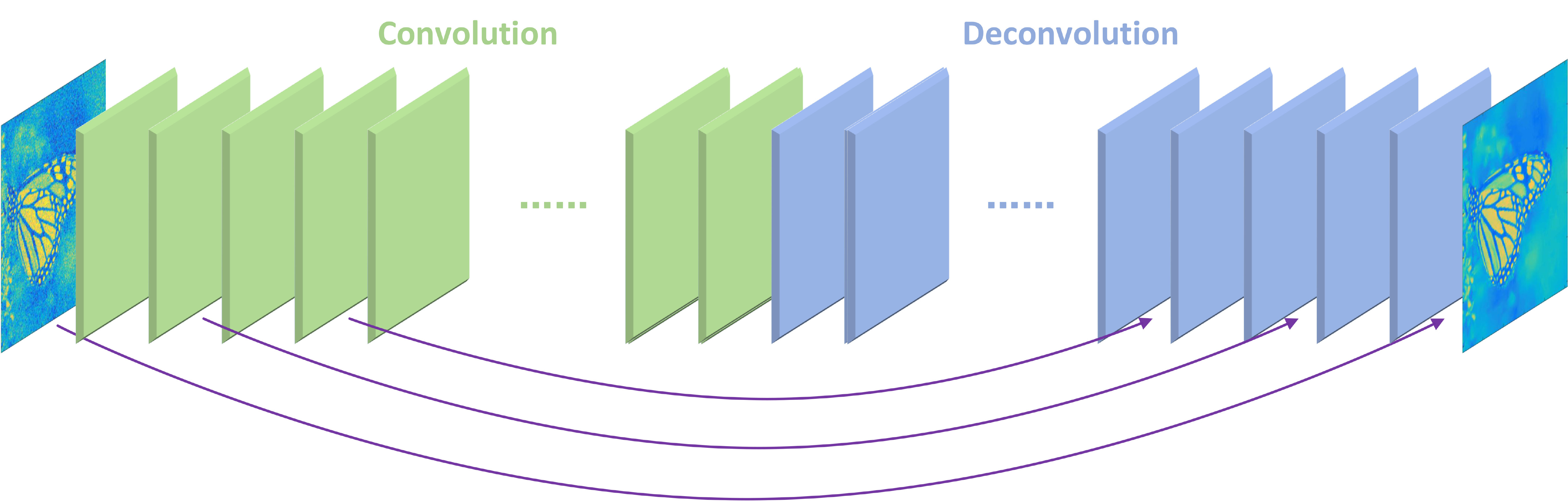}
\caption{The overall architecture of our proposed network. The network
contains layers of symmetric convolution (encoder) and deconvolution (decoder).
Skip shortcuts are connected every a few (in our experiments, two) layers from
convolutional feature maps to their mirrored deconvolutional feature maps.
The response from a convolutional layer is directly propagated to the corresponding
mirrored deconvolutional layer, both forwardly and backwardly.}
\label{fig1}
\end{figure*}

\section{Very deep convolutional auto-encoder for image restoration}
\label{sec:main}

The proposed framework mainly contains a chain of convolutional layers and symmetric
deconvolutional layers, as shown in Figure \ref{fig1}. Skip connections are connected
symmetrically from convolutional layers to deconvolutional layers. We term our method
``RED-Net''---very deep Residual Encoder-Decoder Networks.

\subsection{Architecture}

The framework is fully convolutional (and deconvolutional.  Deconvolution is essentially unsampling convolution). Rectification layers are added
after each convolution and deconvolution. For low-level image restoration problems, we
use neither pooling nor unpooling in the network as usually pooling discards useful image
details that are essential for these tasks. It is worth mentioning that since the convolutional
and deconvolutional layers are symmetric, the network is essentially pixel-wise prediction,
thus the size of input image can be arbitrary. The input and output of the network are images
of the same size $w\times h\times c$, where $w$, $h$ and $c$ are width, height and number of channels.

Our main idea is that the convolutional layers act as a feature extractor, which preserve the
primary components of objects in the image and meanwhile eliminating the corruptions.
After forwarding through the convolutional layers, the corrupted input  image is converted into
a ``clean" one. The subtle details of the image contents may be lost during this process.
The deconvolutional layers are then combined to recover the details of image contents.
The output of the deconvolutional layers is the recovered clean version of the input image.
Moreover, we add skip connections  from a convolutional layer to its corresponding
mirrored deconvolutional layer. The passed convolutional feature maps are summed to the
deconvolutional feature maps element-wise, and passed to the next layer after rectification.
Deriving from the above architecture, we have used two networksvin our experiments, which are of 20 layers
 and 30 layers
respectively, for image denoising, image super-resolution, JPEG deblocking and image inpainting.

\subsection{Deconvolution decoder}

Architectures combining layers of convolution and deconvolution~\cite{DBLP:conf/iccv/NohHH15,
hong2015decoupled} have been proposed for semantic segmentation recently. In contrast to
convolutional layers, in which multiple input activations within a filter window are fused
to output a single activation, deconvolutional layers associate a single input activation with
multiple outputs. Deconvolution is usually used as {\em learnable up-sampling layers}.

 In our network,
the convolutional layers successively down-sample the input image content into a  small
size abstraction. Deconvolutional layers then up-sample the abstraction back into its original resolution.

Besides the use of skip connections, a main difference between our model and
~\cite{DBLP:conf/iccv/NohHH15,hong2015decoupled} is that our network is fully convolutional and
deconvolutional, i.e., without pooling and un-pooling. The reason is that for low-level image restoration,
the aim is to eliminate low level corruption while preserving image details instead of learning
image abstractions. Different from high-level applications such as segmentation or recognition,
pooling typically eliminates the abundant image details and can deteriorate restoration performance.

\begin{figure}[htb!]
\centering
\includegraphics[width=0.48\textwidth]{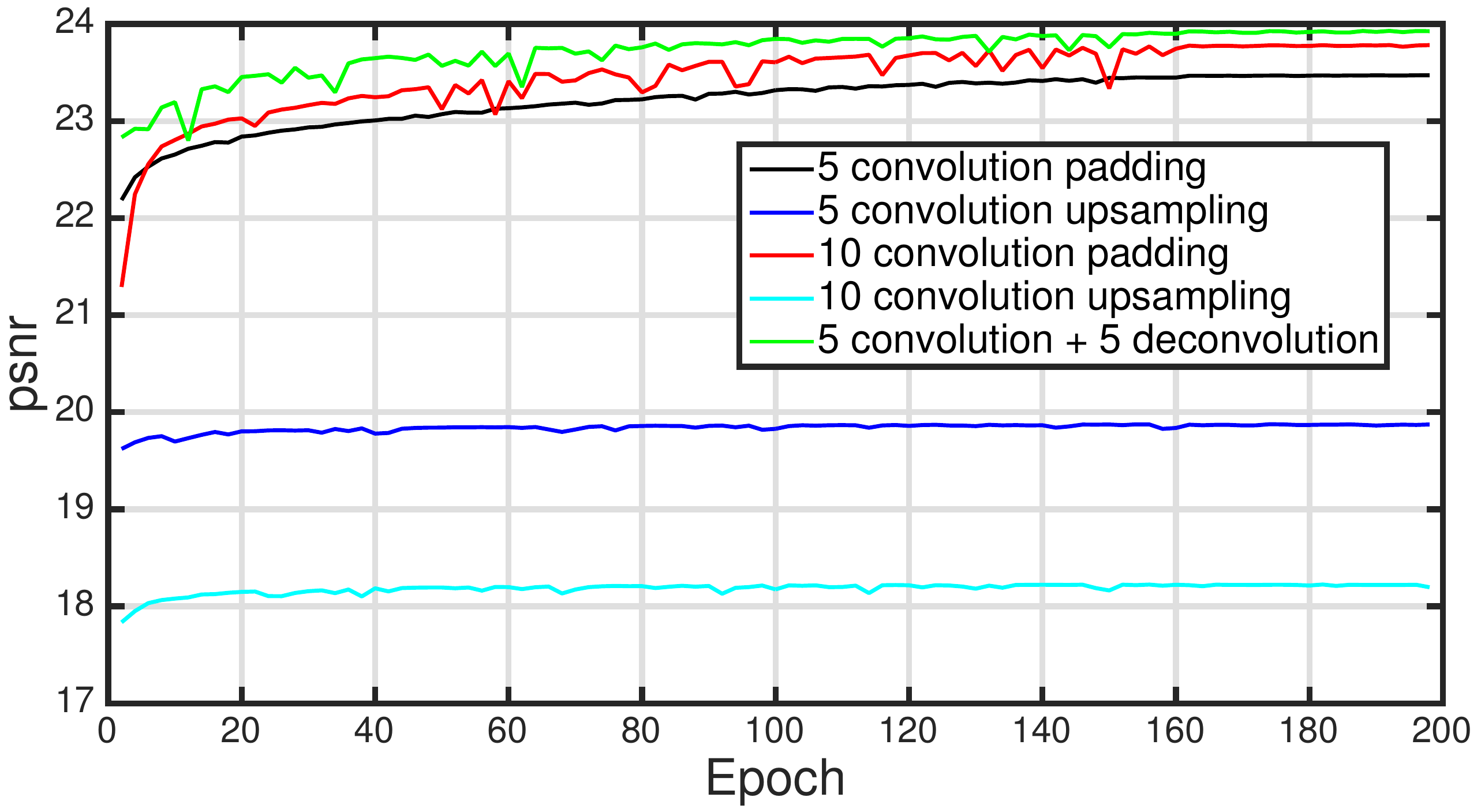}
\caption{ PSNR  values  on the validation set during training. Our model  exhibits better PSNR
than the compared ones upon convergence.}
\label{fig2}
\end{figure}

One can simply replace deconvolution with convolution, which results in an architecture that is
very similar to recently proposed very deep fully convolutional neural networks
~\cite{DBLP:conf/cvpr/LongSD15,DBLP:journals/pami/DongLHT16}. However, there exist essential
differences between a fully convolution model and our model. Take image denoising as an example.
We compare the 5-layer and 10-layer fully convolutional network with our network
(combining convolution and deconvolution, but without skip connection). For fully convolutional
networks, we use padding or up-sampling the input to make the input and output be of the same size.
For our network, the first 5 layers are convolutional and the second 5 layers are deconvolutional.
All the other parameters for training are identical, i.e., trained with SGD and learning rate of
$10^{-6}$, noise level $\sigma=70$. The Peak Signal-to-Noise Ratio (PSNR) on the validation set
is reported, which shows that using deconvolution works better than the fully convolutional
counterpart, as shown in Figure \ref{fig2}.

\begin{figure}[htb!]
\centering
\subfigure[]{ \includegraphics[width=0.48\textwidth]{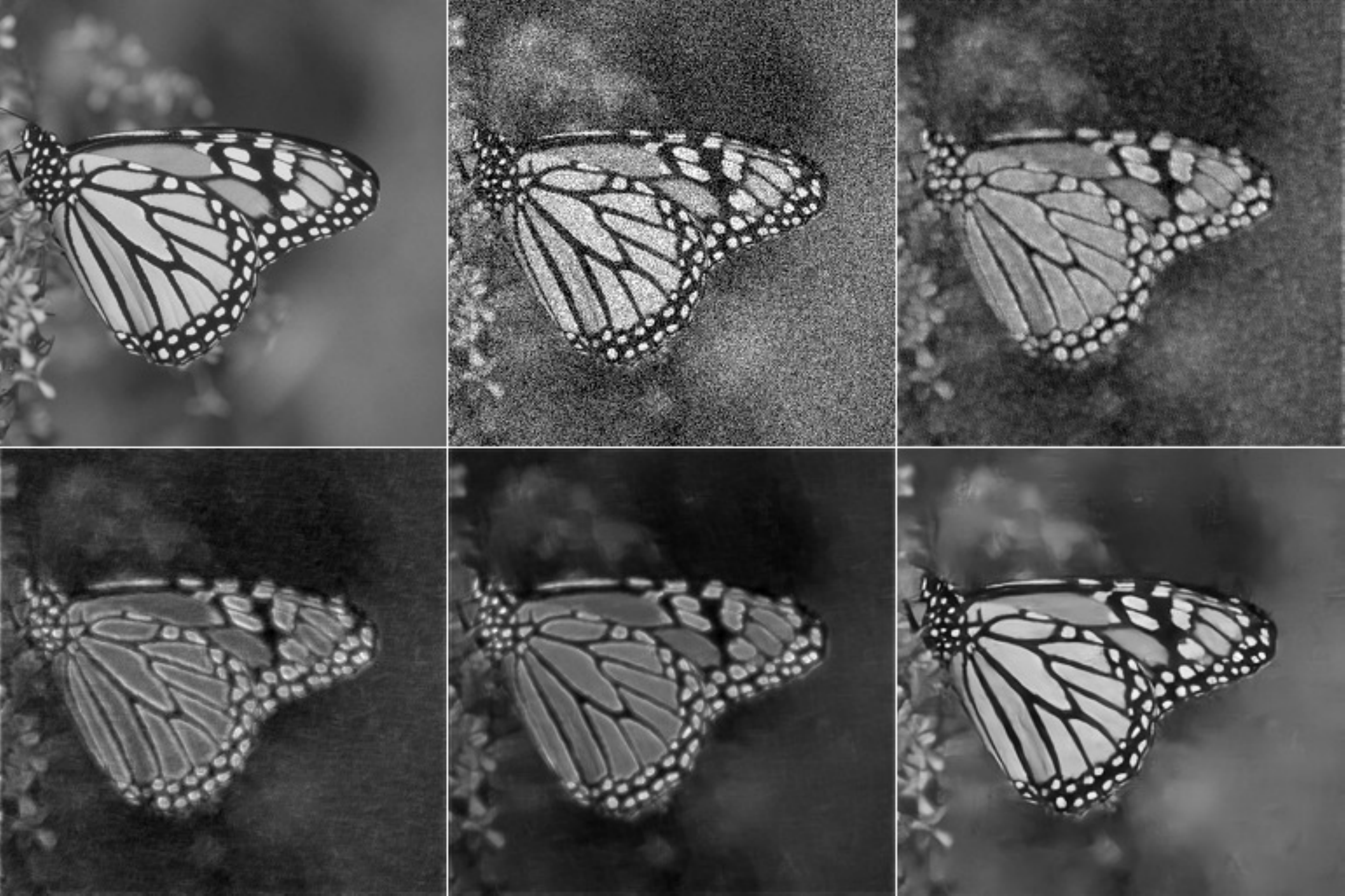} }
\subfigure[]{ \includegraphics[width=0.48\textwidth]{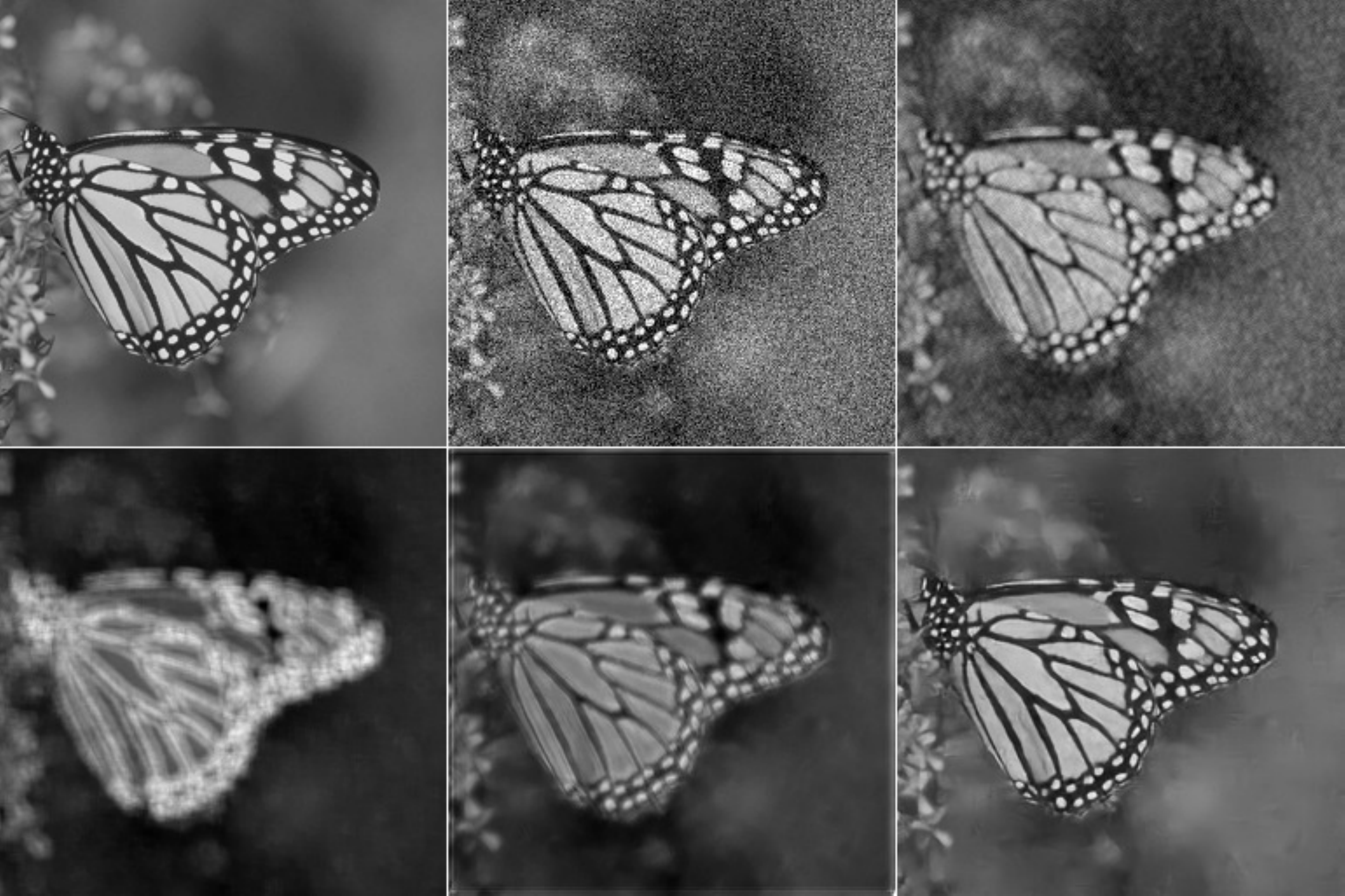} }
\caption{ (a) Visualization of the 10-layer fully convolutional network. The images from
top-left to bottom-right are: clean image, noisy image, output of conv-2, output of conv-5,
output of conv-8 and output of conv-10, where ``conv-$i$" stands for the $i$-th convolutional layer;
(b) Visualization of the 10-layer convolutional and deconvolutional network. The images from
top-left to bottom-right are: clean image, noisy image, output of conv-2, output of conv-5,
output of deconv-3 and output of deconv-5, where ``deconv-$i$" stands for the $i$-th deconvolutional layer.}
\label{fig3}
\end{figure}

Furthermore, in Figure \ref{fig3}, we visualize some results that are outputs of layer 2, 5, 8 and 10
from the 10-layer fully convolutional network and ours. In the fully convolution case, the noise
is eliminated step by step, i.e., the noise level is reduced after each layer. During this process,
the details of the image content may be lost. Nevertheless, in our network, convolution  preserves
the primary image content. Then deconvolution is used to compensate the details.

\subsection{Skip connections}

An intuitive question is that, is a network with deconvolution able to recover image details from
the image abstraction only? We find that in shallow networks with only a few layers
of convolution layers, deconvolution is able to recover the details. However, when the
network goes deeper or using operations such as max pooling, even with deconvolution layers, it does not work
that well, possibly because too much details are already lost in the convolution and pooling.

The second question is that, when our network goes deeper, does it achieve performance gain?
We observe that deeper networks in image restoration tasks tend to easily suffer from
performance degradation. The reason may be two folds. First of all, with more layers of
convolution, a significant amount of image details could be lost or corrupted. Given only the image abstraction,
recovering its details is an under-determined problem. Secondly, in terms of optimization,
deep networks often suffer from gradients vanishing and become much harder to train---a problem
that is well addressed in the literature of neural networks.

\begin{figure}[htb!]
\centering
\includegraphics[width=0.48\textwidth]{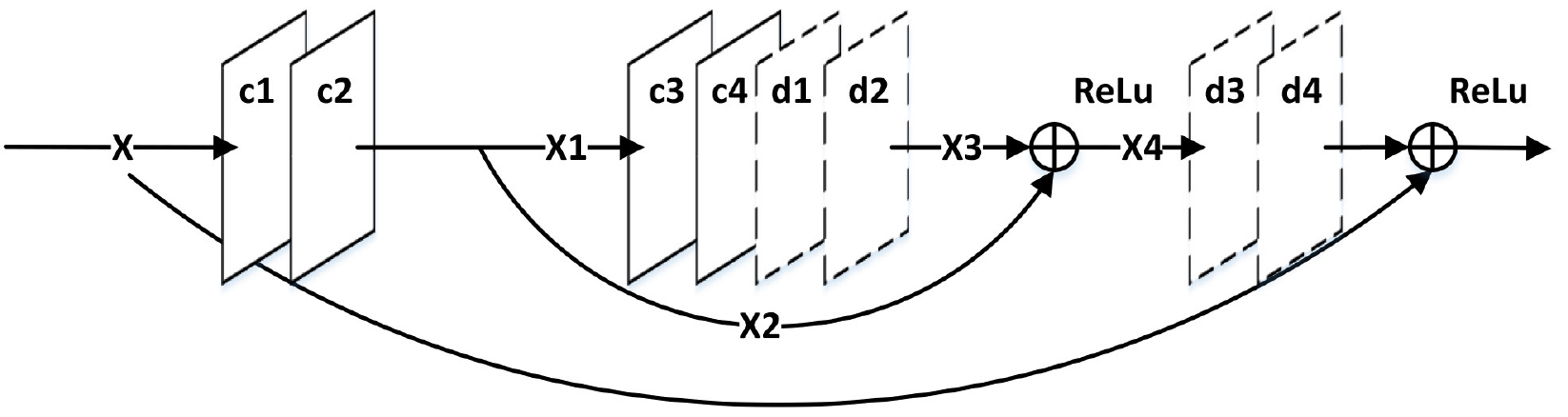}
\caption{An example of a building block in the proposed framework. The rectangle in solid and
dotted lines denote convolution and deconvolution respectively. $\oplus$ denotes element-wise sum of feature maps.}
\label{fig4}
\end{figure}

To address the above two problems, inspired by highway networks \cite{DBLP:journals/corr/SrivastavaGS15}
and deep residual networks \cite{DBLP:journals/corr/HeZRS15}, we add skip connections between
two corresponding convolutional and deconvolutional layers as shown in Figure \ref{fig1}.
A building block is shown in Figure \ref{fig4}. There are two reasons for using such connections.
First, when the network goes deeper, as mentioned above, image details can be lost, making deconvolution
weaker in recovering them. However, the feature maps passed by skip connections carry much image detail,
which helps deconvolution to recover an improved clean version of the image. Second, the skip connections also achieve
benefits on back-propagating the gradient to bottom layers, which makes training deeper network much
easier as observed in \cite{DBLP:journals/corr/SrivastavaGS15} and \cite{DBLP:journals/corr/HeZRS15}.

Note that our skip layer connections are very different from the ones proposed in
\cite{DBLP:journals/corr/SrivastavaGS15} and \cite{DBLP:journals/corr/HeZRS15}, where the only concern
is on the optimization side. In our case, we want to pass information of the convolutional feature maps
to the corresponding deconvolutional layers. The very deep highway networks
\cite{DBLP:journals/corr/SrivastavaGS15} are essentially feedforward long short-term memory (LSTMs)
with forget gates, and the CNN layers of deep residual network \cite{DBLP:journals/corr/HeZRS15}
are feedforward LSTMs without gates. Note that our networks are in general not in the format of
standard feedforward LSTMs.

Instead of directly learning the mappings from the input $X$ to the output $Y$, we would like the network
to fit the residual~\cite{DBLP:journals/corr/HeZRS15} of the problem, which is denoted as $\mathcal{F}(X)=Y-X$.
Such a learning strategy is applied to inner blocks of the encoding-decoding network to make training more
effective. Skip connections are passed every two convolutional layers to their mirrored deconvolutional
layers. Other configurations are possible and our experiments show that this configuration already works
very well. Using such shortcuts makes the network easier to be trained and gains restoration performance
by increasing the network depth.

\subsection{Training}

In general, there are three types of layers in our network: convolution, deconvolution
and element-wise sum. Each layer is followed by a Rectified Linear Unit (ReLU)
~\cite{DBLP:conf/icml/NairH10}. Let $X$ be the input, the convolutional and
deconvolutional layers are expressed as:
\begin{equation}
F(X) = \max(0,W_k * X + B_k),
\end{equation}
where $W_k$ and $B_k$ represent the filters and biases, and $*$ denotes either
convolution or deconvolution operation for the convenience of formulation.
For element-wise sum layer, the output is the element-wise sum of two inputs
of the same size, followed by the ReLU activation:
\begin{equation}
F(X_1,X_2) = \max(0, X_1 + X_2)
\end{equation}

Learning the end-to-end mapping from corrupted images to clean images needs to
estimate the weights $\Theta$ represented by the convolutional and deconvolutional
kernels. Specifically, given a collection of $N$ training sample pairs $\{X^i,Y^i\}$,
where $X^i$ is a noisy image and $Y^i$ is the clean version as the groundtruth.
We minimize the following Mean Squared Error (MSE):
\begin{equation}
  \mathcal{L}(\Theta) = \frac{1}{N}\sum_{i=1}^{N}\|\mathcal{F}(X^i;\Theta)-Y^i\|_F^2.
\label{eq1}
\end{equation}

Traditionally, a  network can learn the mapping from the corrupted image to the clean version
directly. However, our network learns for the additive corruption from the input since there
is a skip connection between the input and the output of the network.
We found that optimizing for the corruption converges better than
optimizing for the clean image. In the extreme case, if the input is a clean image, it would be easier
to push the network to be zero mapping (learning the corruption) than to fit an identity
mapping (learning the clean image) with a stack of nonlinear layers.

We implement and train our network using Caffe~\cite{jia2014caffe}. Empirically, we find
that using Adam~\cite{DBLP:journals/corr/KingmaB14} with base learning rate of $10^{-4}$ for
training converges faster than traditional stochastic gradient descent (SGD). The base
learning rate for all layers are the same, different from ~\cite{DBLP:journals/pami/DongLHT16,
DBLP:conf/nips/JainS08}, in which a smaller learning rate is set for the last layer.
This  is not necessary in our network. Specifically, gradients with respect to the
parameters of $i$th layer is firstly computed as:
\begin{equation}
g = \nabla_{\theta_i}\mathcal{L}(\theta_i).
\end{equation}
Then, the two momentum vectors are computed as:
\begin{equation}
m = \beta_1m + (1 - \beta_1)g,\quad v = \beta_2v + (1-\beta_2)g^2.
\end{equation}
The update rule is:
\begin{equation}
\alpha = \alpha\sqrt{1-\beta_2^t}/(1-\beta_1^t), \quad \theta_i=\theta_i-\alpha m/(\sqrt{v}+\epsilon).
\end{equation}
$\beta_1$, $\beta_2$ and $\epsilon$ are set as the recommended values in~\cite{DBLP:journals/corr/KingmaB14}.

300 images from the Berkeley Segmentation Dataset (BSD)~\cite{MartinFTM01} are used to
generate image patches as the training set for each image restoration task.

\subsection{Testing}

Although trained on local patches, our network can perform restoration on images of arbitrary sizes.
Given a testing image, one can simply go forward through the network, which is already able to
 outperform existing methods. To achieve even better results, we propose
to process a corrupted image on multiple orientations. Different from segmentation, the
filter kernels in our network only eliminate the corruptions, which is usually not sensitive
to the orientation of image contents in low level restoration tasks. Therefore, we can rotate
and mirror flip the kernels and perform forward multiple times, and then average the output to
achieve an ensemble of multiple tests. We see that this can lead to slightly better performance.

\section{Discussions}

\label{sec:disc}

\subsection{Analysis on the architecture}

Assume that we have a network with $L$ layers, and skip connections are passed every
layer in the first half of the network. For the convenience of presentation, we
denote $F_c$ and $F_d$  the convolution and deconvolution operation in each layer
and do not use ReLU. According to the architecture described in the last section,
we can obtain the output of the $i$-th layer as follows:
\begin{equation}
   X_i = \left\{
   \begin{array}{ll}
   X_{L-i} + F_d(X_{i-1}), & i\geq L/2;\\
   F_c(X_{i-1}).           & i<L/2.
   \end{array} \right.
\label{eq5}
\end{equation}
It is easy to observe that our skip connections indicate identity mapping.
The output of the network is:
\begin{equation}
X_L = X_0 + F_d(X_{L-1}).
\end{equation}
Recursively, we can compute $X_L$ more specifically as follows according to Equation \eqref{eq5}:
\begin{equation}
\begin{aligned}
X_L & = X_0 + F_d(X_{L-1}) \\
    & = X_0 + F_d(X_1 + F_d(X_{L-2})) \\
    & = X_0 + F_d(X_1) + F_d^2(X_2+F_d(X_{L-3})) \\
    & ......   \\
    & = X_0 + F_d(X_1) + F_d^2(X_2) + ... +F_d^{L/2-1}(X_{L/2-1}) \\
    & + F_d^{L/2}(X_{L/2}).
\end{aligned}
\label{eq6}
\end{equation}
Since $F_d^{L/2}(X_{L/2})$ can be expressed as $F_d^{L/2}(F_c^{L/2}(X_0))$, we convert Equation \eqref{eq6} as:
\begin{equation}
X_L = F_d^{L/2}(F_c^{L/2}(X_0)) + \sum_{i=0}^{L/2-1} F_d^i(X_i).
\label{eq7}
\end{equation}
In Equation \eqref{eq7}, the term $F_d^{L/2}(F_c^{L/2}(X_0))$ is actually the output
of the given network without skip connections. The difference here is that by adopting
the skip connection, we decode each feature maps $X_i, 0\leq i <L/2$ in the first half
network and integrate them to the output. The most significant benefit is that they carry
important image details, which helps to reconstruct clean image. Moreover, the term
$\sum_{i=0}^{L/2-1} F_d^i(X_i)$ indicates that these details are represented at
different levels. It is intuitive  to see the following fact. It may not be easy to tell what information
 is needed for
reconstructing clean images using only one feature maps encoding the image abstraction;
but much easier if there are multiple feature maps encoding different levels of image abstraction.

\subsection{Gradient back-propagation}

For back-propagation, a layer receives gradients from the layers that it is connected to.
As an example shown in Figure \ref{fig4}, $X$ is the input of the first layer,
after two convolutional layers $c1$ and $c2$, the output is $X_1$. To update the parameters
represented as $\theta_2$ of $c2$, we compute the derivative of $\mathcal{L}$ with
respect to $\theta_2$ as follows:
\begin{equation}
\nabla_{\theta_2}\mathcal{L}(\theta_2) = \frac{\partial\mathcal{L}}{\partial X_1}\frac{\partial X_1}{\partial\theta_2} + \frac{\partial\mathcal{L}}{\partial X_2}\frac{\partial X_2}{\partial\theta_2}
\label{eq2}
\end{equation}
where using $X_1$ and $X_2$ is only for the clarity of presentation, they are essentially
the same. We can further formulate \eqref{eq2} as:
\begin{equation}
\nabla_{\theta_2}\mathcal{L}(\theta_2) = \frac{\partial\mathcal{L}}{\partial X_4}\frac{\partial X_4}{\partial X_3}\frac{\partial X_3}{\partial X_1}\frac{\partial X_1}{\partial\theta_2} + \frac{\partial\mathcal{L}}{\partial X_4}\frac{\partial X_4}{\partial X_2}\frac{\partial X_2}{\partial\theta_2}.
\label{eq3}
\end{equation}
Only $\frac{\partial\mathcal{L}}{\partial X_4}\frac{\partial X_4}{\partial X_3}\frac{\partial X_3}{\partial X_1}\frac{\partial X_1}{\partial\theta_2}$ is computed if we do not use skip connections, and its magnitide
 may become very
small after back-propagating through many layers from the top in very deep networks.
However, $\frac{\partial\mathcal{L}}{\partial X_4}\frac{\partial X_4}{\partial X_2}\frac{\partial X_2}{\partial\theta_2}$
carries larger gradients since it does not have to go through layers of $d2$, $d1$, $c4$ and $c3$ in this example.
Thus with the first term only, it is more unlikely to approach zero grdients.
 As we can see, the skip connection helps to update
the filters in bottoms layers, and thus makes training easier.

\subsection{Training with symmetric skip connections}

The aim of restoration is to eliminate corruption while preserving the image details
as mush as possible. Previous works typically use shallow networks for low-level image
restoration tasks. The reason may be that deeper networks can destroy the image details,
which is undesired for pixel-wise dense regression. Even worse, using very deep networks
may easily suffer from training issues such as gradient vanishing. Using skip
connections in a very deep network can address both  of the above two problems.

Firstly, we design experiments to show that using skip connections is beneficial for
image detail presering. Specifically, two networks are trained for image denoising
with a noise level of $\sigma=70$.

\begin{figure}[htb!]
\centering
\includegraphics[width=0.48\textwidth]{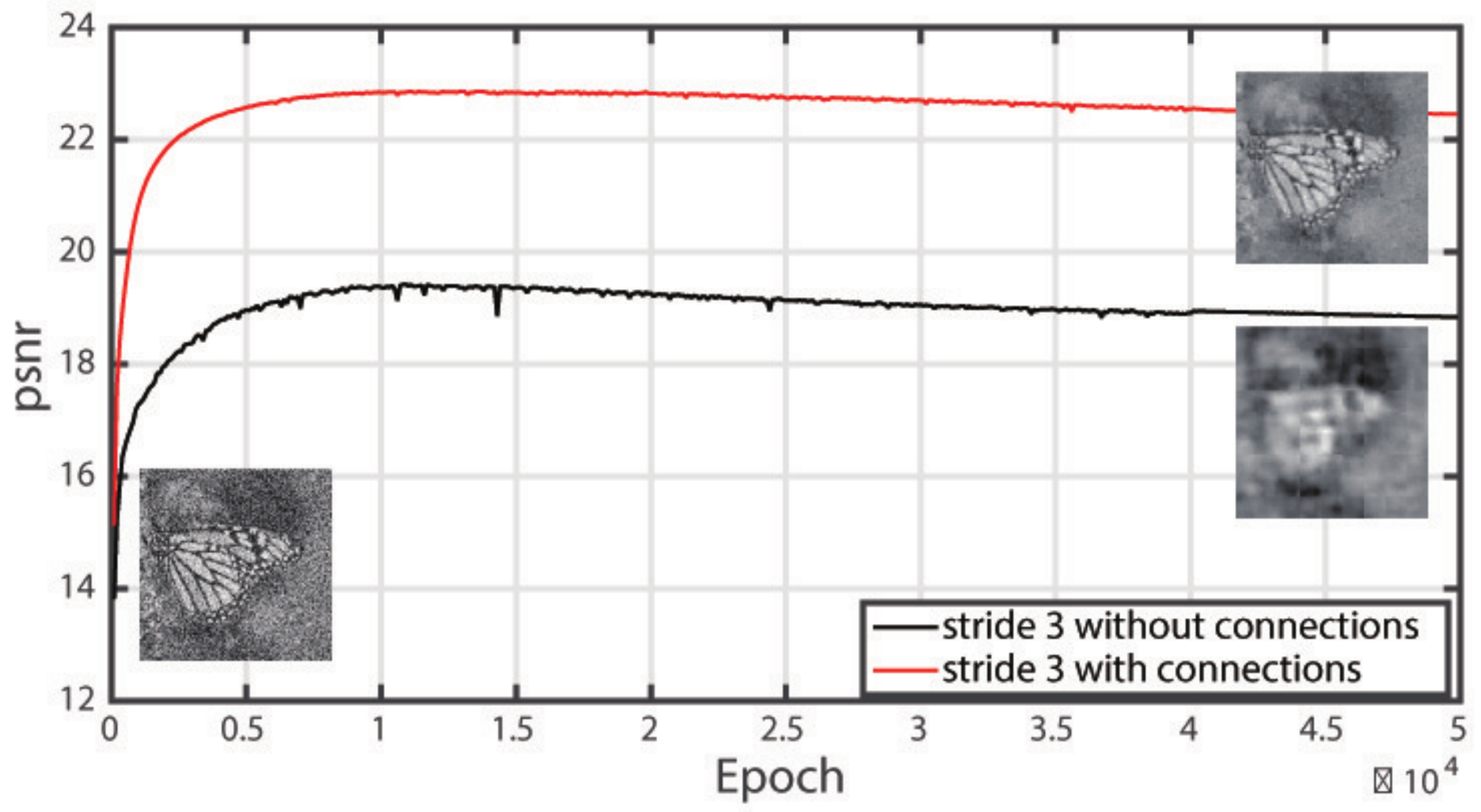}
\caption{Recovering image details using deconvolution and skip connections. Skip
connections are beneficial in recovering image details.}
\label{fig5}
\end{figure}

(a) In the first network, we use 5 layers of $3\times3$ convolution with stride 3.
The input size of training data is $243\times243$, which results in a vector after
5 layers of convolution, encoding the very high level abstraction of the image. Then
deconvolution is used to recover the input from the feature vector. The results are
shown in Figure \ref{fig5}. We can observe that it is challenging for deconvolution to recover
details from only a vector encoding the abstraction of the input. This phenomenon
implies that  simply using deep networks for image restoration may not lead to satisfactory results.

(b) The second network uses the same settings as the first one, but adding
skip connections. The results are show in Figure \ref{fig5}. Compared to the first
network, the one with skip connections can recover the input and achieves much
better PSNR values. This is easy to understand since the feature maps with abundant
details at bottom layers are directly passed to the top layers.

\begin{figure}[b!]
\centering
\includegraphics[width=0.48\textwidth]{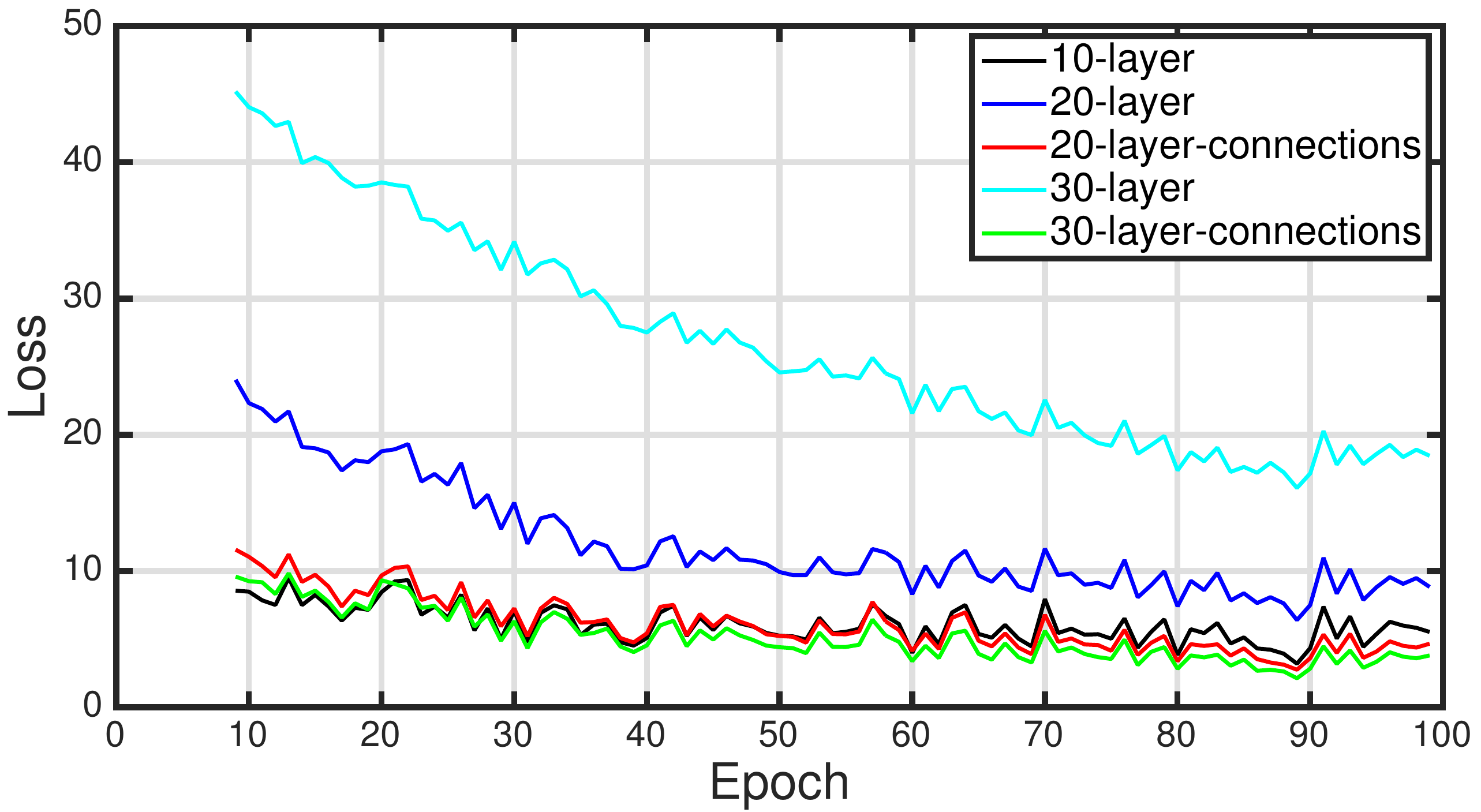}
\caption{The training loss on the training set during training.}
\label{fig6}
\end{figure}

Secondly, we train and compare five different networks to show that using skip
connections help to back-propagate gradient in training to better fit the end-to-end
mapping, as shown in Figure \ref{fig6}. The five networks are: 10, 20 and 30 layer
networks without skip connections;  and 20, 30 layer networks with skip connections.
As  can be seen, the training loss increases when the network going deeper without
shortcuts (similar phenomenon is also observed in \cite{DBLP:journals/corr/HeZRS15}).
On the validation set, deeper networks without shortcuts achieve lower PSNR and we
even observe over-fitting for the 30-layer network. These results may be due to  the
gradient vanishing problem. However, we obtain smaller training errors on the training set and
higher PSNR and better generalization capability on the testing set when using skip connections.

\subsection{Comparison with the deep residual network \cite{DBLP:journals/corr/HeZRS15}}

\begin{figure}[t!]
\centering
\includegraphics[width=0.48\textwidth]{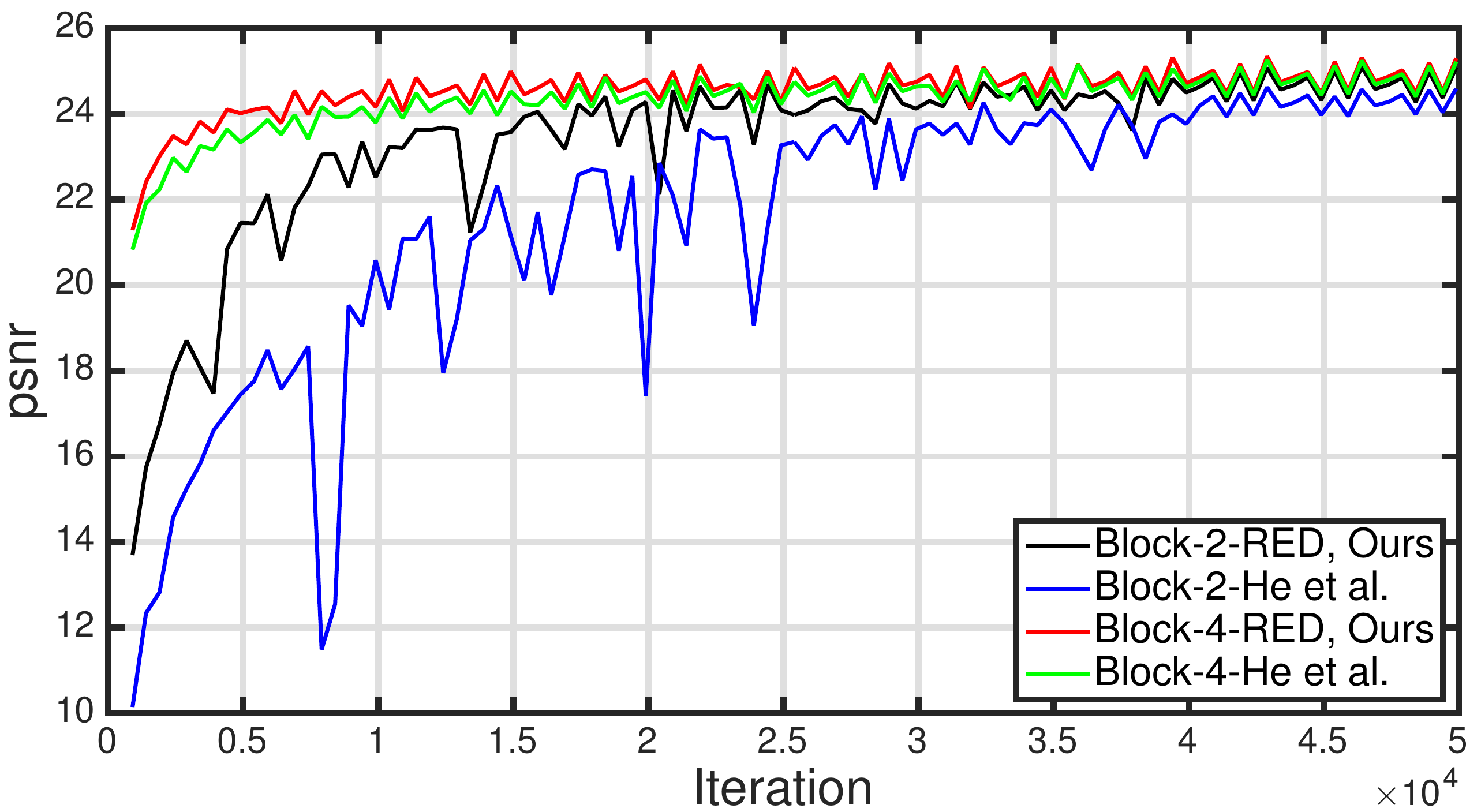}
\caption{Comparisons of skip connections in ~\cite{DBLP:journals/corr/HeZRS15} and our
model, where ``Block-$i$-RED" is the connections in our model with block size $i$ and
``Block-$i$-He et al." is the connections in He et al.~\cite{DBLP:journals/corr/HeZRS15}
with block size $i$; The PSNR values on the validation set during training: the PSNR at
the last iteration for the curves are: 25.08, 24.59, 25.30 and 25.21.}
\label{fig8}
\end{figure}

One may use different types of skip connections in our network. A straightforward
alternate is that in ~\cite{DBLP:journals/corr/HeZRS15}. In ~\cite{DBLP:journals/corr/HeZRS15},
 skip connections are added to divide the network into sequential blocks. A benefit
of our model is that our skip connections have element-wise correspondence, which can be
very important in pixel-wise prediction problems such image denoising. We carry out
experiments to compare these two types of skip connections. Here the block size indicates
the span of the connections. The results are shown in Figure \ref{fig8}. We can observe
that our connections often converge to a better optimum, demonstrating that element-wise
correspondence can be important. Meanwhile, our long range skip connections pass the image
detail directly from bottom layers to top layers. If we use the skip connection type in
~\cite{DBLP:journals/corr/HeZRS15}, the network may still lose some image details.

\subsection{Testing efficiency}

To apply deep learning models on devices with limited computing power such as mobile
phones, one has to speed-up the testing phase. For our network, we propose to use
down-sampling in convolutional layers to reduce the size of the feature maps. In order
to obtain an output of the same size as the input, deconvolution is used to up-sample
the feature maps in the symmetric deconvolutional layers.  Thus, the testing
efficiency can be well improved with almost negligible  performance degradation.

\begin{figure}[t!]
\centering
\includegraphics[width=0.48\textwidth]{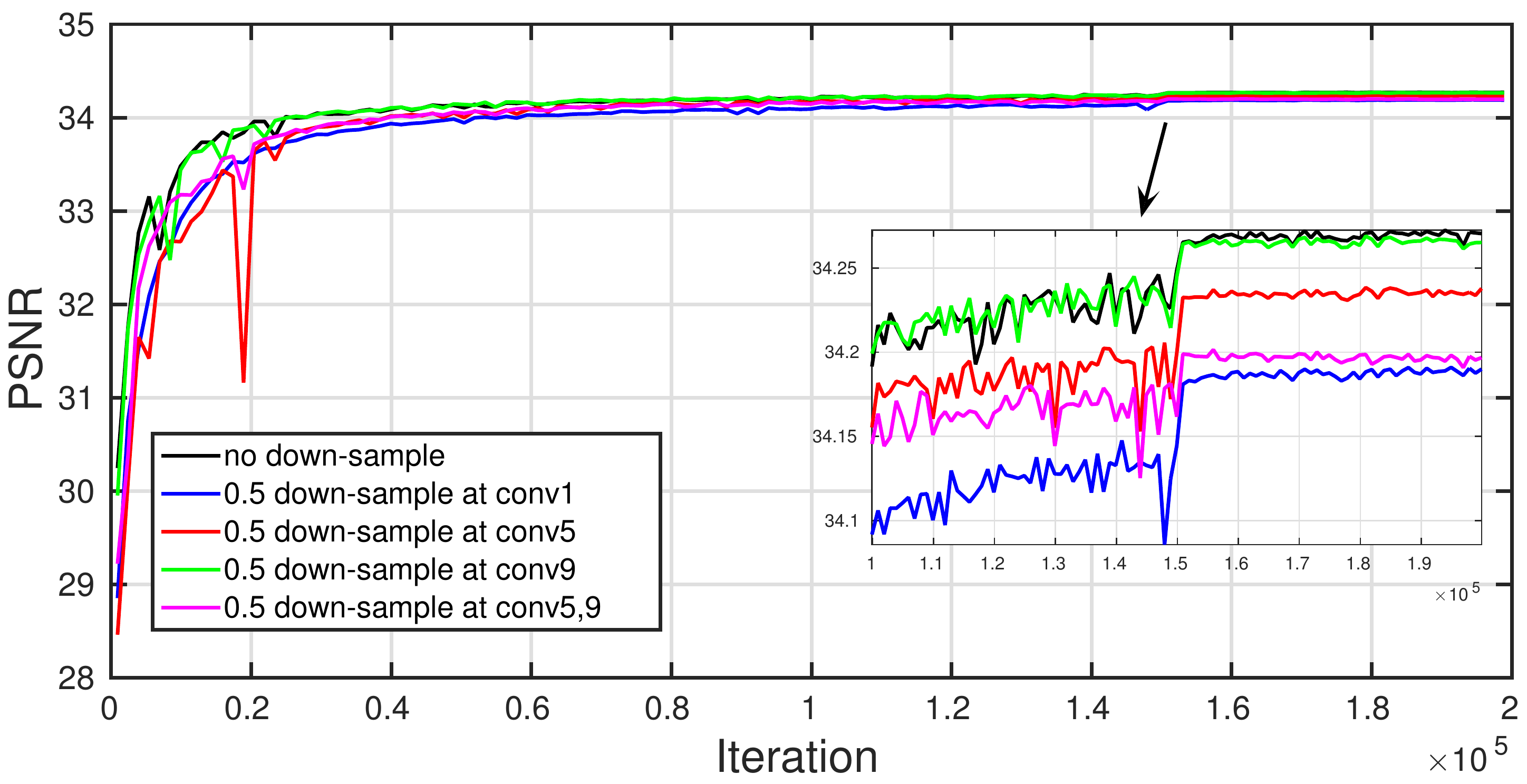}
\caption{The PSNRs on the validation set with different down-sampling strategies.
``down-sample at conv-i" denotes that down-sampling is used in the $i$th convolutional
layer, and up-sampling is used in its symmetric deconvolutional layer.}
\label{fig16}
\end{figure}

In specific, we use stride $=$ 2 in convolutional layers to down-sample the feature maps.
Down-sampling at different convolutional layers are tested on image denoising, as shown
in Figure \ref{fig16}. We test an image of size 160$\times$240 on an i7-2600 CPU, the testing
time for ``no down-sample", ``down-sample at conv1", ``down-sample at conv5",
``down-sample at conv9", "down-sample at conv5,9" are 3.17s, 0.84s, 1.43s, 2.00s and
1.17s respectively.

The main observation is that the testing PSNRs may slightly degrade according to the scale
reduction of the feature map in the entire network. The down-sampling in the first
convolutional layer reduces the size of the feature maps to 1/4, which leads to alomst 4x
faster in testing, but the PSNR only degrades less than 0.1 compared to the network without
down-sampling. The down-sampling in "conv9" reduces 1/3 of the testing time, but the
performance is almost as well as that without down-sampling. As a result, an "earlier"
down-sampling may lead to slightly worse performance, but it achieves much faster testing
efficiency. It should be a trade-off in different application situations.

\section{Experiments}
\label{sec:exp}

In this section, we first provide some experimental results and analysis
on different parameters, including filter number, filter size, training patch
size and skip connection step size, of the network.

Then, evaluation of image
restoration tasks including image denoising, image super-resolution, JPEG image
deblocking, non-blind image debluring and image inpainting are conducted and
compared against a few existing state-of-the-art methods in each topic.
Peak Signal-to-Noise Ratio (PSNR) and Structural SIMilarity (SSIM) index are
calculated for evaluation. For our method, which is denoted as RED-Net,
we implement three versions: RED10 contains 5 convolutional and deconvolutional
layers without shortcuts, RED20 contains 10 convolutional and deconvolutional
layers with shortcuts of step size 2, and RED30 contains 15 convolutional
and deconvolutional layers with shortcuts of step size 2.

\subsection{Network parameters}

Although we have observed that deeper networks tend to achieve better image restoration
performance, there exist more problems related to different parameters to be investigated.
We carried out image denoising experiments on three folds: (a) filter number,
(b) filter size, (c) training patch size and (d) step size of skip connections, to
show the effects of different parameters.

\begin{figure}[t!]
\centering
\includegraphics[width=0.48\textwidth]{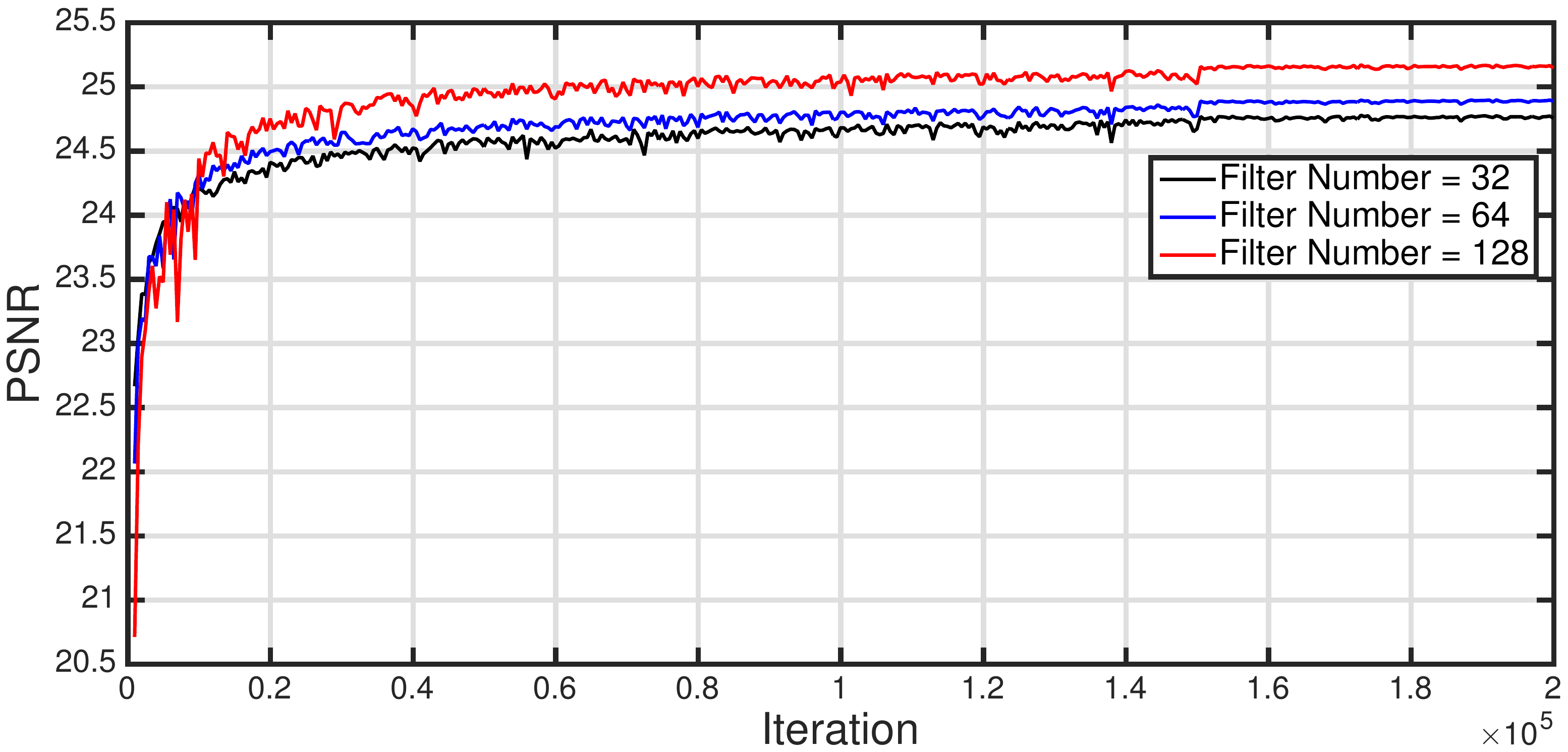}
\caption{The PSNR values on the validation set during training with different number of filters.}
\label{fig9}
\end{figure}

For different filter numbers, we fix the filter size as $3\times3$, training patch size
as 50$\times$50 and skip connection step size as 2. Different filter numbers of 32, 64
and 128 are tested, and the PSNR values recorded on the validation set during training
are shown in Figure \ref{fig9}. To converge, the training iterations for different number
of filters are similar, but better optimum can be obtained with more filters. However,
a smaller number of filters is preferred if a fast testing speed is desired.

For the experiments on filter size, we set the filter number to be 64, training patch
size as 50$\times$50, skip connection step size as 2.

\begin{figure}[t!]
\centering
\includegraphics[width=0.48\textwidth]{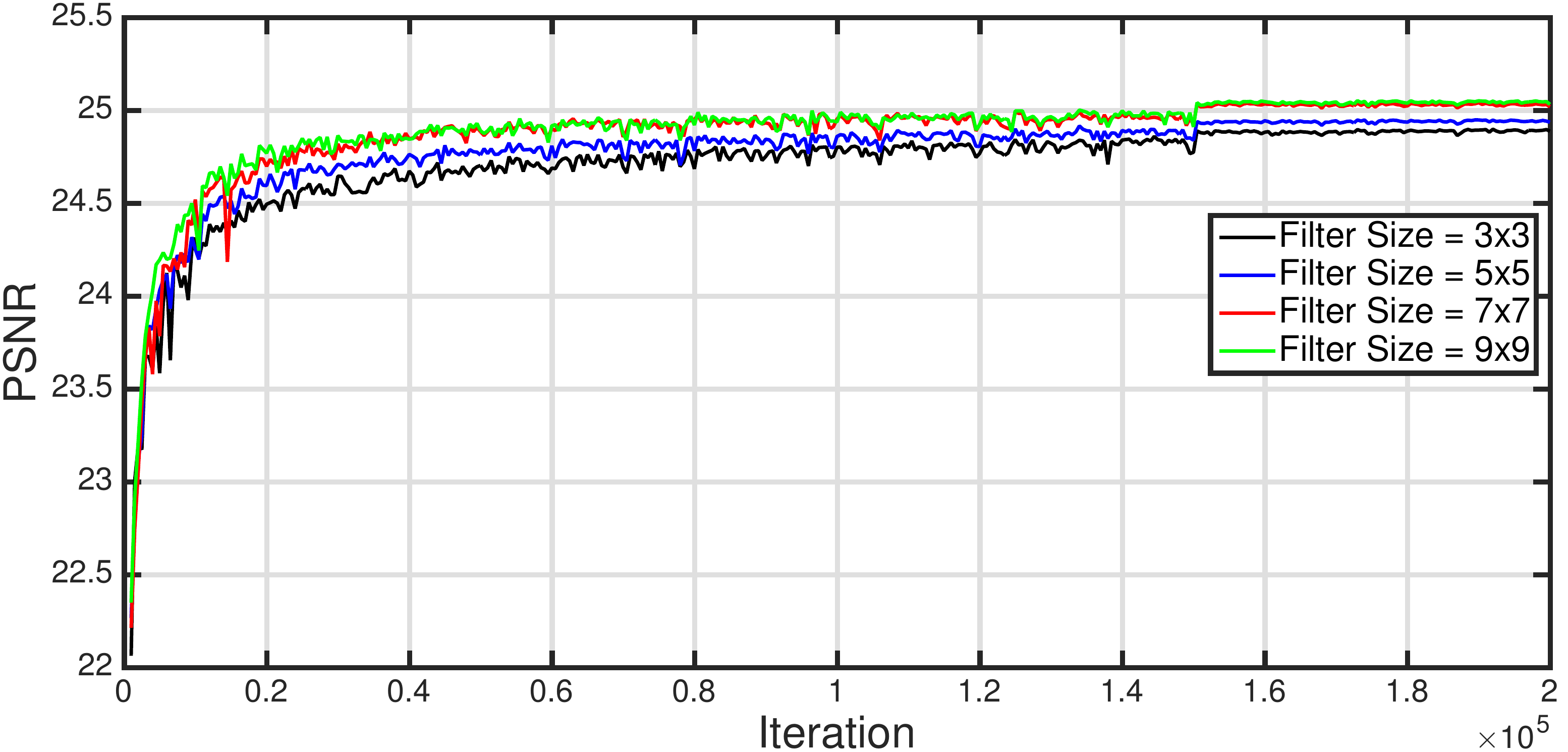}
\caption{The PSNR values on the validation set during training with different size of filters.}
\label{fig10}
\end{figure}

Filter size of 3$\times$3,
5$\times$5, 7$\times$7, 9$\times$9 are tested. Figure \ref{fig10} show the PSNR values
on the validation set while training. It is clear that larger filter size leads to
better performance. Different from high-level tasks~\cite{DBLP:conf/eccv/ZeilerF14,
DBLP:journals/corr/SermanetEZMFL13,DBLP:journals/corr/SimonyanZ14a} which favor smaller
filter sizes, larger filter size tends to obtain better performance in low-level image
restoration applications.

However, there  may  exist a bottle neck  as  the performance
of 9$\times$9 is almost as the same as 7$\times$7 in our experiments. The reason may be
that for high-level tasks, the networks have to learn image abstraction for classification, which is usually
very different from the input pixels. Larger filter size may result in larger respective fields,
but also made the networks more difficult to train and converge to a poor optimum. Using
smaller filter size is mainly beneficial for convergence in such complex mappings.

In contrast, for low-level image restoration, the training is not as difficult as that in high-level
applications since only a bias is needed to be learned to revise the corrupted pixel.
In this situation, utilizing neighborhood information in the mapping stage is more important,
since the desired value for a pixel should be predicted from its neighbor pixels.
However, using larger filter size inevitably increases the complexity (e.g., filter
size of 9$\times$9 is 9 times more complex as 3$\times$3) and training time.

\begin{figure}[b!]
\centering
\includegraphics[width=0.48\textwidth]{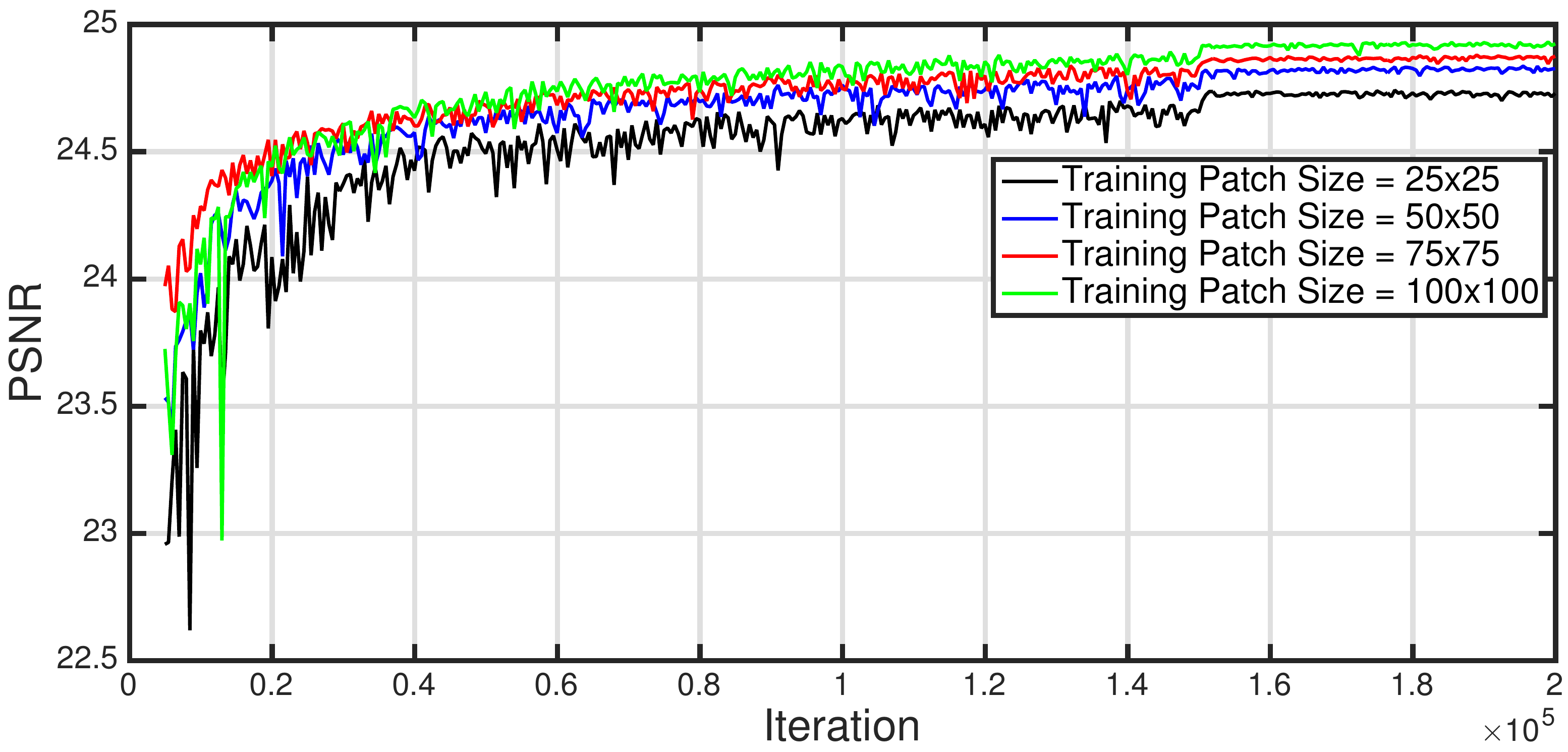}
\caption{The PSNR values on the validation set during training with different size of training patch.}
\label{fig11}
\end{figure}

For the training patch size, we set the filter number to be 64, filter size as 3$\times$3,
skip connection step size as 2. Then we test different training patch sizes of 25$\times$25,
50$\times$50, 75$\times$75, 100$\times$100, as shown in Figure \ref{fig11}.

Better
performance is achieved with larger training patch size. The reason can be two folds.
First of all, since the network essentially performs pixel-wise prediction, if the number
of training patches are the same, larger size of training patch results in more pixels
to be used, which is equivalent to using more training data. Secondly, the corruptions
in image restoration tasks can be described as some types of latent distributions. Larger
size of training patch contains more pixels that better capture the latent distributions
to be learned, which consequently helps the network to fit the corruptions better.

As we can see, the ``width" of the network is as crucial as the ``depth" in training
a network with satisfactory image restoration performance. However, one should always make a
trade-off between the performance and speed.

\begin{figure}[htb!]
\centering
\includegraphics[width=0.48\textwidth]{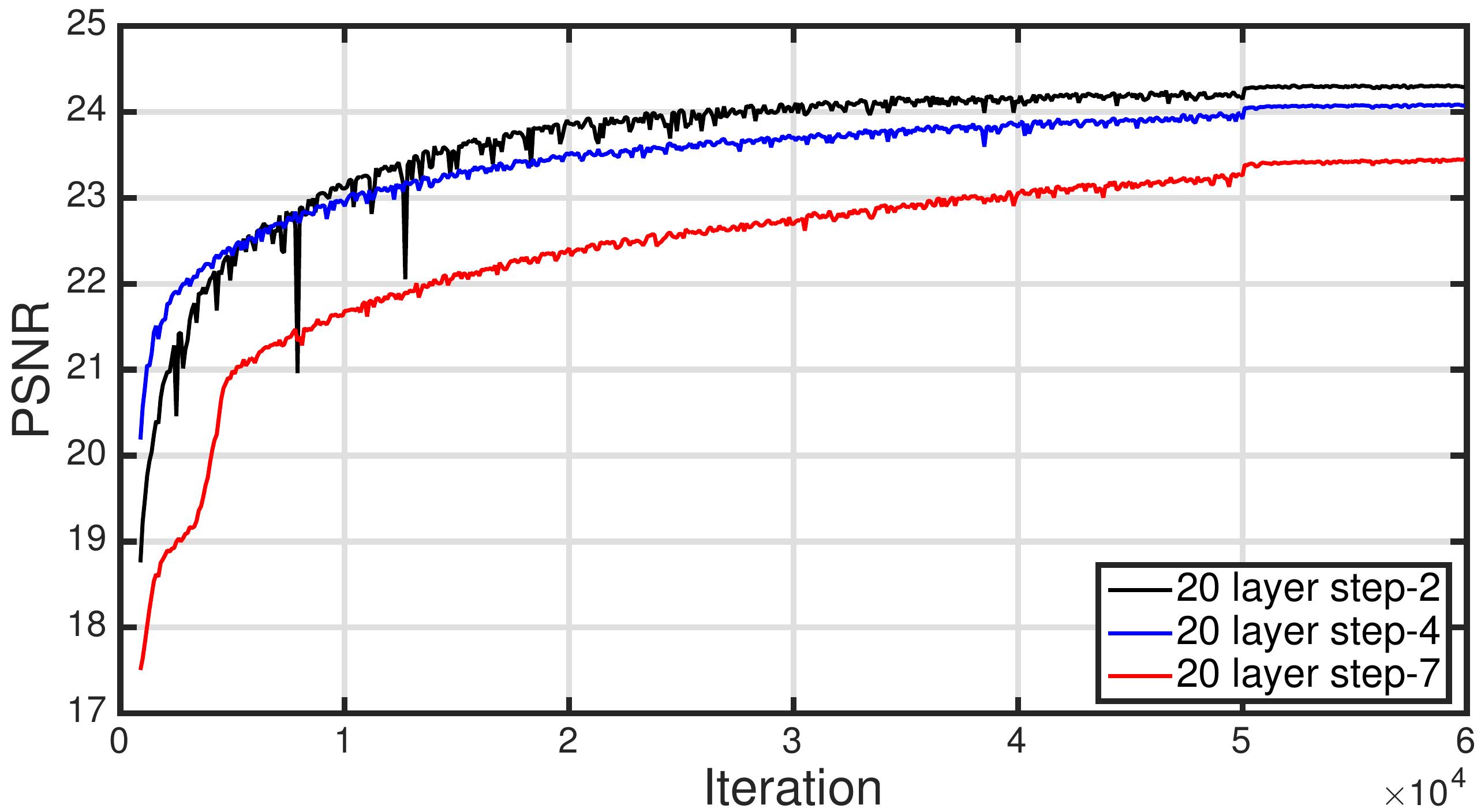}
\caption{The PSNR values on the validation set during training.}
\label{fig7}
\end{figure}

We also provide the experiments of different step sizes of shortcuts, as shown in
Figure \ref{fig7}. A smaller step size of shortcuts achieves better performance than
a larger one. We believe that a smaller step size of shortcuts makes it easier
to back-propagate the gradient to bottom layers, thus tackle the gradient vanishing
issue better. Meanwhile, a small step size of shortcuts essentially passes more direct information.

\subsection{Image denoising}

\begin{figure}
\centering
\includegraphics[width=0.48\textwidth]{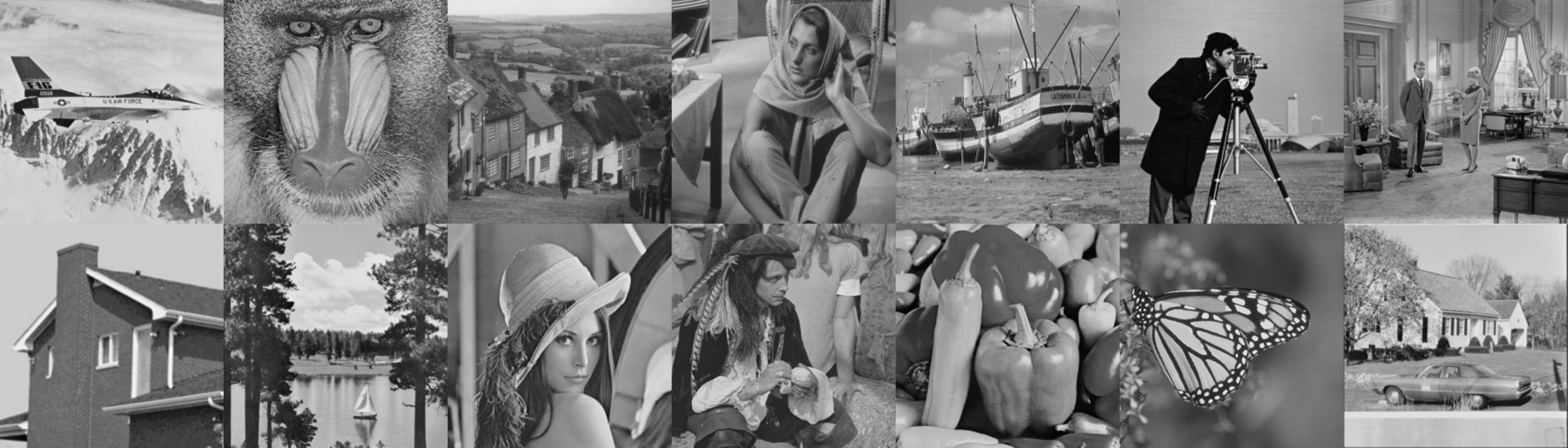}
\caption{The 14 testing images for denoising.}
\label{fig12}
\end{figure}

Image denoising experiments are performed on two datasets: 14 common benchmark images
~\cite{DBLP:conf/iccv/XuZZZF15,DBLP:conf/iccv/ChenZY15,DBLP:conf/cvpr/LiuXZG15,
DBLP:conf/cvpr/GuZZF14}, as show in Figure \ref{fig12} and the BSD dataset.

As a common
experimental setting in the literature, additive Gaussian noises with zero mean and
standard deviation $\sigma$ are added to the image to test the performance of denoising
methods. In this paper we test noise level $\sigma$ of 10, 30, 50 and 70. BM3D
~\cite{DBLP:journals/tip/DabovFKE07}, NCSR~\cite{DBLP:journals/tip/DongZSL13}, EPLL
~\cite{DBLP:conf/iccv/ZoranW11}, PCLR~\cite{DBLP:conf/iccv/ChenZY15}, PGPD~\cite{DBLP:conf/iccv/XuZZZF15}
and WMMN~\cite{DBLP:conf/cvpr/GuZZF14} are compared with our method. For these methods,
we use the source code released by their authors and test on the images with their default parameters.

{\bf{Evaluation on the 14 images}}  Table \ref{table2} presents the PSNR and SSIM
results of $\sigma$ 10, 30, 50, and 70. We can make some observations from the results.
First of all, the 10 layer convolutional and deconvolutional network has already achieved
better results than the state-of-the-art methods, which demonstrates that combining
convolution and deconvolution for denoising works well, even without any skip connections.

Moreover, when the network goes deeper, the skip connections proposed in this paper help to
achieve even better denoising performance, which exceeds the existing best method WNNM
~\cite{DBLP:conf/cvpr/GuZZF14} by 0.32dB, 0.43dB, 0.49dB and 0.51dB on noise levels of $\sigma$
being 10, 30, 50 and 70 respectively. While WNNM is only slightly better than the second
best existing method PCLR~\cite{DBLP:conf/iccv/ChenZY15} by 0.01dB, 0.06dB, 0.03dB and 0.01dB
respectively, which shows the large improvement of our model.

Last, we can observe that the
more complex the noise is, the more improvement our model achieves than other methods.
Similar observations can be made on the evaluation of SSIM.

\begin{table*}[htb!]
\centering
\caption{Average PSNR and SSIM results of $\sigma$ 10, 30, 50, 70 on 14 images.}
\begin{tabular}{c|c c c c c c c c c} \hline
              &\multicolumn{9}{c}{PSNR}            \\ \hline
              &BM3D   &EPLL   &NCSR   &PCLR   &PGPD   &WNNM   &RED10  &RED20  &RED30          \\ \hline
  $\sigma=10$ &34.18  &33.98  &34.27  &34.48  &34.22  &34.49  &34.62  &34.74  &\textbf{34.81} \\ \hline
  $\sigma=30$ &28.49  &28.35  &28.44  &28.68  &28.55  &28.74  &28.95  &29.10  &\textbf{29.17} \\ \hline
  $\sigma=50$ &26.08  &25.97  &25.93  &26.29  &26.19  &26.32  &26.51  &26.72  &\textbf{26.81} \\ \hline
  $\sigma=70$ &24.65  &24.47  &24.36  &24.79  &24.71  &24.80  &24.97  &25.23  &\textbf{25.31} \\ \hline
              &\multicolumn{9}{c}{SSIM}            \\ \hline
  $\sigma=10$ &0.9339 &0.9332 &0.9342 &0.9366 &0.9309 &0.9363 &0.9374 &0.9392 &\textbf{0.9402} \\ \hline
  $\sigma=30$ &0.8204 &0.8200 &0.8203 &0.8263 &0.8199 &0.8273 &0.8327 &0.8396 &\textbf{0.8423} \\ \hline
  $\sigma=50$ &0.7427 &0.7354 &0.7415 &0.7538 &0.7442 &0.7517 &0.7571 &0.7689 &\textbf{0.7733} \\ \hline
  $\sigma=70$ &0.6882 &0.6712 &0.6871 &0.6997 &0.6913 &0.6975 &0.7012 &0.7177 &\textbf{0.7206} \\ \hline
\end{tabular}
\label{table2}
\end{table*}

{\bf{Evaluation on BSD200}} For the BSD dataset, 300 images are used for training and
the remaining 200 images are used for denoising to show more experimental results.
For efficiency, we convert the images to gray-scale and resize them to smaller images.
Then all the methods are run on the dataset to get average PSNR and SSIM results of
$\sigma$ 10, 30, 50, and 70, as shown in Table \ref{table3}. For existing methods,
their denoising performance does not differ much, while our model achieves 0.38dB,
0.47dB, 0.49dB and 0.42dB higher of PSNR over WNNM \cite{DBLP:conf/cvpr/GuZZF14}.

\begin{table*}[htb!]
\centering
\caption{Average PSNR and SSIM results of $\sigma$ 10, 30, 50, 70 on BSD.}
\begin{tabular}{ c|c c c c c c c c c } \hline
			  &\multicolumn{9}{c}{PSNR}            \\ \hline
              &BM3D   &EPLL   &NCSR   &PCLR   &PGPD   &WNNM   &RED10  &RED20  &RED30 \\ \hline
  $\sigma=10$ &33.01  &33.01  &33.09  &33.30  &33.02  &33.25  &33.49  &33.59  &\textbf{33.63} \\ \hline
  $\sigma=30$ &27.31  &27.38  &27.23  &27.54  &27.33  &27.48  &27.79  &27.90  &\textbf{27.95} \\ \hline
  $\sigma=50$ &25.06  &25.17  &24.95  &25.30  &25.18  &25.26  &25.54  &25.67  &\textbf{25.75} \\ \hline
  $\sigma=70$ &23.82  &23.81  &23.58  &23.94  &23.89  &23.95  &24.13  &24.33  &\textbf{24.37} \\ \hline
              &\multicolumn{9}{c}{SSIM}            \\ \hline
  $\sigma=10$ &0.9218 &0.9255 &0.9226 &0.9261 &0.9176 &0.9244 &0.9290 &0.9310 &\textbf{0.9319} \\ \hline
  $\sigma=30$ &0.7755 &0.7825 &0.7738 &0.7827 &0.7717 &0.7807 &0.7918 &0.7993 &\textbf{0.8019} \\ \hline
  $\sigma=50$ &0.6831 &0.6870 &0.6777 &0.6947 &0.6841 &0.6928 &0.7032 &0.7117 &\textbf{0.7167} \\ \hline
  $\sigma=70$ &0.6240 &0.6168 &0.6166 &0.6336 &0.6245 &0.6346 &0.6367 &0.6521 &\textbf{0.6551} \\ \hline
\end{tabular}
\label{table3}
\end{table*}

{\bf{Blind denoising}} We also perform blind denoising to show the superior performance
of our network. In blind denoising, the training set consists of image patches of different
levels of noises, and a 30-layer network is trained on this training set. In the testing
phase, we test noisy images with $\sigma$ of 10, 30, 50 and 70 using this model. The evaluation
results are shown in Table \ref{table4}. Although training with different levels of corruption,
we can observe that the performance of our network degrades comparing to the case in which
using separate models for denoising. This is reasonable because the network has to fit much
more complex mappings. However, it still beats the existing methods. For PSNR evaluation,
our blind denoising model achieves the same performance as WNNM  \cite{DBLP:conf/cvpr/GuZZF14}
on $\sigma=10$, and outperforms
WNNM \cite{DBLP:conf/cvpr/GuZZF14} by
 0.35dB, 0.43dB and 0.40dB on $\sigma$ = 30, 50 and 70 respectively, which is still
marginal improvements. For SSIM evaluation, our network is 0.0005, 0.0141, 0.0199 and 0.0182
higher than WNNM. The performance improvement is more obvious on BSD dataset. The 30-layer
network outperforms the second best method WNNM \cite{DBLP:conf/cvpr/GuZZF14} by
 0.13dB, 0.4dB, 0.43dB, 0.41dB for
PSNR and 0.0036, 0.0173, 0.0191, 0.0198 for SSIM.

\begin{table}[b!]
\centering
\caption{Average PSNR and SSIM results for image denoising using a single 30-layer network.}
\begin{tabular}{ c|c c c c } \hline
       &\multicolumn{4}{c}{14 images}                     \\ \hline
       &$\sigma=10$ &$\sigma=30$ &$\sigma=50$ &$\sigma=70$ \\ \hline
  PSNR &34.49       &29.09       &26.75       &25.20       \\ \hline
  SSIM &0.9368      &0.8414      &0.7716      &0.7157      \\ \hline
     &\multicolumn{4}{c}{BSD200}                          \\ \hline
       &$\sigma=10$ &$\sigma=30$ &$\sigma=50$ &$\sigma=70$ \\ \hline
  PSNR &33.38       &27.88       &25.69       &24.36       \\ \hline
  SSIM &0.9280      &0.7980      &0.7119      &0.6544      \\ \hline
\end{tabular}
\label{table4}
\end{table}

\begin{figure*}
\centering
\subfigure{\includegraphics[width=1\textwidth]{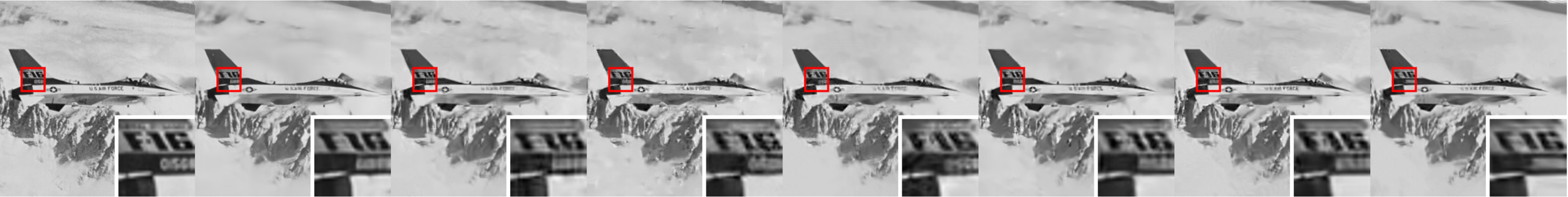} }
\subfigure{\includegraphics[width=1\textwidth]{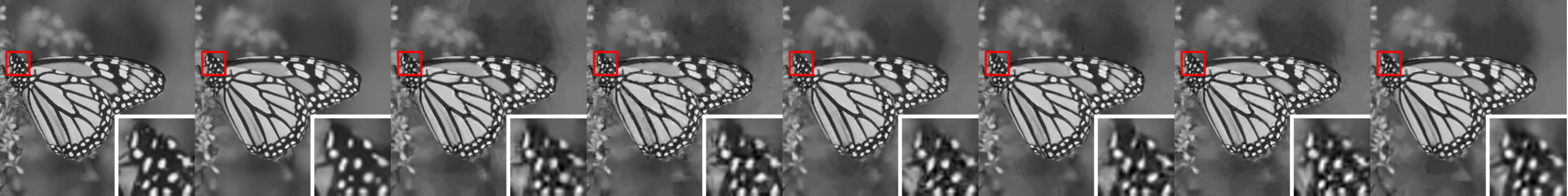} }
\subfigure{\includegraphics[width=1\textwidth]{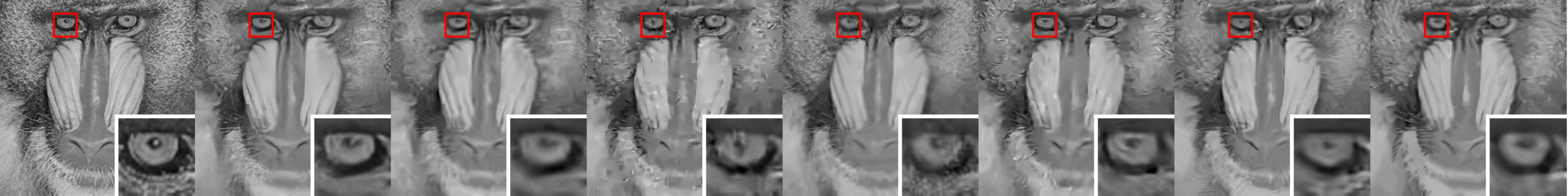} }
\subfigure{\includegraphics[width=1\textwidth]{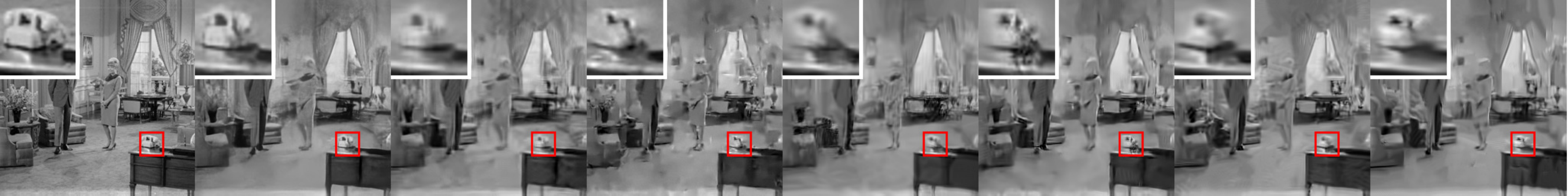} }
\caption{Visual results of image denoising. Images from left to right column are:
clean image; the recovered image of RED30, BM3D, EPLL, NCSR, PCLR, PGPD, WNNM.}
\label{fig13}
\end{figure*}

{\bf{Visual results}} Some visual results are shown in Figure \ref{fig13}. We highlight
some details of the clean image and the recovered ones by different methods. The first
observation is that our method better recovers the image details, as we can see from the
third and fourth rows, which is due to the high PSNR we achieve by minimizing the pixel-wise
Euclidean loss.

 Moreover, we can observe from the first and second rows that our network
obtains more visually smooth results than other methods. This may due to the testing
strategy which average the output of different orientations.

\subsection{Image super-resolution}

For super-resolution, The high-resolution image is first down-sampled with scaling
factor parameters of 2, 3 and 4 respectively. Since the size of the input and output of our
network are the same, we up-sample the low-resolution image to its original size as
the input of our network. We compare our network with SRCNN~\cite{DBLP:journals/pami/DongLHT16},
NBSRF~\cite{DBLP:conf/iccv/SalvadorP15}, CSCN~\cite{DBLP:conf/iccv/WangLYHH15},
CSC~\cite{DBLP:conf/iccv/GuZXMFZ15}, TSE~\cite{DBLP:conf/cvpr/HuangSA15} and ARFL+
~\cite{DBLP:conf/cvpr/SchulterLB15} on three dataset: Set5, Set14 and BSD100.

The results of the compared methods are either cited from their original papers or obtained
using the released source code by the authors.

{\bf{Evaluation on Set 5}} The evaluation on Set5 is shown in Table \ref{table5}.
In general, our 10-layer network already outperforms the compared methods, and we
achieve better performance with deeper networks.

 The second best method is CSCN, which
is also a recently proposed neural network based method. Compared to CSCN, our
30-layer network exceeds it by 0.52dB, 0.56dB, 0.47dB on PSNR and 0.0032, 0.0063,
0.0094 on SSIM respectively.

The larger scaling parameter is, the better improvement
our method can make, which demonstrates that our network is better at fitting
complex corruptions than other methods.

\begin{table*}[htb!]
\centering
\caption{Average PSNR and SSIM results of scaling 2, 3 and 4 on Set5.}
\begin{tabular}{ c|c c c c c c c c c }  \hline
              &\multicolumn{9}{c }{PSNR}            \\ \hline
           &SRCNN  &NBSRF  &CSCN   &CSC    &TSE    &ARFL+   &RED10    &RED20   &RED30           \\ \hline
  $s = 2$  &36.66  &36.76  &37.14  &36.62  &36.50  &36.89   &37.43    &37.62   &\textbf{37.66}  \\ \hline
  $s = 3$  &32.75  &32.75  &33.26  &32.66  &32.62  &32.72   &33.43    &33.80   &\textbf{33.82}  \\ \hline
  $s = 4$  &30.49  &30.44  &31.04  &30.36  &30.33  &30.35   &31.12    &31.40   &\textbf{31.51}  \\ \hline
              &\multicolumn{9}{c }{SSIM}            \\ \hline
  $s = 2$  &0.9542 &0.9552 &0.9567 &0.9549 &0.9537 &0.9559  &0.9590   &0.9597  &\textbf{0.9599} \\ \hline
  $s = 3$  &0.9090 &0.9104 &0.9167 &0.9098 &0.9094 &0.9094  &0.9197   &0.9229  &\textbf{0.9230} \\ \hline
  $s = 4$  &0.8628 &0.8632 &0.8775 &0.8607 &0.8623 &0.8583  &0.8794   &0.8847  &\textbf{0.8869} \\ \hline
\end{tabular}
\label{table5}
\end{table*}

{\bf{Evaluation on Set 14}} The evaluation on Set14 is shown in Table \ref{table6}.
The improvement on Set14 in not as significant as that on Set5, but we can still
observe that the 30-layer network achieves higher PSNR and SSIM than the second
best CSCN for 0.23dB, 0.06dB, 0.1dB and 0.0049, 0.0070, 0.0098. The performance
on 10-layer, 20-layer and 30-layer RED-Net also does not improve that much as on
Set5, which may imply that Set14 is more difficult to perform image super-resolution.

\begin{table*}[htb!]
\centering
\caption{Average PSNR and SSIM results of scaling 2, 3 and 4 on Set14.}
\begin{tabular}{c|c c c c c c c c c}  \hline
              &\multicolumn{9}{c}{PSNR}            \\ \hline
           &SRCNN  &NBSRF  &CSCN   &CSC    &TSE    &ARFL+   &RED10    &RED20   &RED30           \\ \hline
  $s = 2$  &32.45  &32.45  &32.71  &32.31  &32.23  &32.52   &32.77    &32.87   &\textbf{32.94}  \\ \hline
  $s = 3$  &29.30  &29.25  &29.55  &29.15  &29.16  &29.23   &29.42    &29.61   &\textbf{29.61}  \\ \hline
  $s = 4$  &27.50  &27.42  &27.76  &27.30  &27.40  &27.41   &27.58    &27.80   &\textbf{27.86}  \\ \hline
              &\multicolumn{9}{c}{SSIM}            \\ \hline
  $s = 2$  &0.9067 &0.9071 &0.9095 &0.9070 &0.9036 &0.9074 &0.9125    &0.9138  &\textbf{0.9144} \\ \hline
  $s = 3$  &0.8215 &0.8212 &0.8271 &0.8208 &0.8197 &0.8201 &0.8318    &0.8343  &\textbf{0.8341} \\ \hline
  $s = 4$  &0.7513 &0.7511 &0.7620 &0.7499 &0.7518 &0.7483 &0.7654    &0.7697  &\textbf{0.7718} \\ \hline
\end{tabular}
\label{table6}
\end{table*}

{\bf{Evaluation on BSD 100}} We also evaluate super-resolution results on BSD100,
as shown in Table \ref{table7}. The overall results are very similar than those on Set5.
CSCN is still the second best method and outperforms other compared methods by large
margin, but its performance is not as good as our 10-layer network. Our deeper networks
obtain performance gains. Compared to CSCN, the 30-layer network achieves higher
PSNR for 0.45dB, 0.38dB, 0.29dB and higher SSIM for 0.0066, 0.0084, 0.0099.

\begin{table*}[htb!]
\centering
\caption{Average PSNR and SSIM results of scaling 2, 3 and 4 on BSD100}
\begin{tabular}{c|c c c c c c c c c}  \hline
              &\multicolumn{9}{c}{PSNR}            \\ \hline
           &SRCNN  &NBSRF  &CSCN   &CSC    &TSE    &ARFL+  &RED10  &RED20   &RED30           \\ \hline
  $s = 2$  &31.36  &31.30  &31.54  &31.27  &31.18  &31.35  &31.85  &31.95   &\textbf{31.99}  \\ \hline
  $s = 3$  &28.41  &28.36  &28.58  &28.31  &28.30  &28.36  &28.79  &28.90   &\textbf{28.93}  \\ \hline
  $s = 4$  &26.90  &26.88  &27.11  &26.83  &26.85  &26.86  &27.25  &27.35   &\textbf{27.40}  \\ \hline
              &\multicolumn{9}{c}{SSIM}            \\ \hline
  $s = 2$  &0.8879 &0.8876 &0.8908 &0.8876 &0.8855 &0.8885 &0.8953 &0.8969  &\textbf{0.8974} \\ \hline
  $s = 3$  &0.7863 &0.7856 &0.7910 &0.7853 &0.7843 &0.7851 &0.7975 &0.7993  &\textbf{0.7994} \\ \hline
  $s = 4$  &0.7103 &0.7110 &0.7191 &0.7101 &0.7108 &0.7091 &0.7238 &0.7268  &\textbf{0.7290} \\ \hline
\end{tabular}
\label{table7}
\end{table*}

{\bf{Comparisons with VDSR~\cite{DBLP:journals/corr/KimLL15b}
and DRCN~\cite{DBLP:journals/corr/KimLL15a}}}
Concurrent to our work \cite{NIPS2016Mao}, networks~\cite{DBLP:journals/corr/KimLL15b,DBLP:journals/corr/KimLL15a}
which incorporate residual learning for image super-resolution are proposed.
In \cite{DBLP:journals/corr/KimLL15b}, a fully convolutional network termed  VDSR
is proposed to learn the residual image for image super-resolution.
The loss layer takes three inputs: residual estimate, low-resolution input and
ground truth high-resolution image, and Euclidean loss is computed between the
reconstructed image (the sum of network input and output) and ground truth.
DRCN~\cite{DBLP:journals/corr/KimLL15a} proposed to use a recursive convolutional block,
which does not increase the number of parameters while increasing the depth of the network.
To ease the training, firstly each recursive layer is supervised to reconstruct the
target high-resolution image (HR). The second proposal is to use a skip-connection
from input to the output. During training, the network has $D$ outputs,
in which the $d$th output $y_d = x + {\rm Rec}(H_d)$. $x$ is the input low-resolution image,
${\rm Rec}()$ denotes the reconstruction layer and $H_d$ is the output of $d$th recursive layer.
The final loss includes three parts: (a) the Euclidean loss between the ground truth and each $y_d$;
(b) the Euclidean loss between the ground truth and the weighted sum of all $y_d$; and
(c) the L2 regularization on the network weights. Although skip connections are used in our network,
VDSR and DCRN to perform identity mapping, their differences are significant.

Firstly, {\em both VDSR and DRCN use one path of connections between the input and output,
which actually models the corruptions.}  In VDSR, the network itself is standard fully convolutional.
DRCN uses recursive convolutional layers that lead to multiple losses, which is different from VDSR.
The skip connections in VDSR and DRCN model the super-resolution problem as learning the residual image,
which actually learns the corruption as in image restoration. In other words,
the residual learning is only conducted in the input-output level (low-resolution and high-resolution images)
in VDSR and DRCN. In contrast, {\em  our network uses multiple skip connections that divide the
network into multiple blocks for residual learning}. Secondly, our skip connections pass
image abstraction of different levels from multiple convolutional layers forwardly.
No such information is used in VDSR and DRCN. In VDSR and DRCN,
the skip connection only pass the input image. However, in our network,
different levels of image abstraction are obtained after the convolutional layers,
and they are passed to the deconvolutional layers for reconstruction. At last,
our skip connections help to back-propagate gradients in different layers. In VDSR and DCRN,
the skip connections do not involve in back-propagating gradients since they connect the input and output,
and there are no weights to be updated for the input low-resolution image.
The image super-resolution comparisons of VDSR, DRCN and our network on Set5,
Set14 and BSD100 are provided in Table~\ref{table11}.

\begin{table*}[htb!]
\centering
\caption{Comparisons between RED30 (ours), VDSR \cite{DBLP:journals/corr/KimLL15b}
and DRCN \cite{DBLP:journals/corr/KimLL15a}: Average PSNR and SSIM results of scaling 2, 3 and 4 on Set5, Set14 and BSD100.}
\begin{tabular}{ c|c|c c c }  \hline
Dataset                 &Scale        &VDSR  (PSNR/SSIM)    &DRCN (PSNR/SSIM)    &RED30 (PSNR/SSIM)  \\ \hline
\multirow{3}{*}{Set5}   &$\times$2    &37.53/0.9587       &37.63/0.9588       &\textbf{37.66}/\textbf{0.9599}      \\
                        &$\times$3    &33.66/0.9213       &33.82/0.9226       &\textbf{33.82}/\textbf{0.9230}      \\
                        &$\times$4    &31.35/0.8838       &\textbf{31.53}/0.8854       &31.51/\textbf{0.8869}      \\ \hline\hline
\multirow{3}{*}{Set14}  &$\times$2    &33.03/0.9124       &\textbf{33.04}/0.9118       &32.94/\textbf{0.9144}      \\
                        &$\times$3    &\textbf{29.77}/0.8314       &29.76/0.8311       &29.61/\textbf{0.8341}      \\
                        &$\times$4    &28.01/0.7674       &\textbf{28.02}/0.7670       &27.86/\textbf{0.7718}      \\ \hline\hline
\multirow{3}{*}{BSD100} &$\times$2    &31.90/0.8960       &31.85/0.8942       &\textbf{31.99}/\textbf{0.8974}      \\
                        &$\times$3    &28.82/0.7976       &28.80/0.7963       &\textbf{28.93}/\textbf{0.7994}      \\
                        &$\times$4    &27.29/0.7251       &27.23/0.7233       &\textbf{27.40}/\textbf{0.7290}      \\ \hline
\end{tabular}
\label{table11}
\end{table*}

{\bf{Blind super-resolution}} The results of blind super-resolution are shown in
Table \ref{table8}. Among the compared methods, CSCN can also deal with different
scaling parameters by repeatedly enlarging the image by a smaller scaling factor.

 Our method is
different from CSCN. Given a low-resolution image as input and the output size,
we first up-sample the input image to the desired size, resulting in an image with
poor details. Then the image is fed into our network. The output is an image of the
same size with fine details. The training set consists of image patches of different
scaling parameters and a single model is trained. Except that CSCN works slightly
better on Set 14 with scaling factors 3 and 4, our network  outperforms the existing methods,
showing that our network works much better in image super-resolution even using only
one single model to deal with complex corruptions.

\begin{table}[htb!]
\centering
\caption{Average PSNR and SSIM results for image super-resolution using a single 30 layer network.}
\begin{tabular}{c|c c c } \hline
       &\multicolumn{3}{c}{Set5}     \\ \hline
       &$s = 2$  &$s = 3$  &$s = 4$  \\ \hline
  PSNR &37.56    &33.70    &31.33    \\ \hline
  SSIM &0.9595   &0.9222   &0.8847   \\ \hline
       &\multicolumn{3}{c}{Set14}    \\ \hline
       &$s = 2$  &$s = 3$  &$s = 4$  \\ \hline
  PSNR &32.81    &29.50    &27.72    \\ \hline
  SSIM &0.9135   &0.8334   &0.7698   \\ \hline
       &\multicolumn{3}{c}{BSD100}   \\ \hline
       &$s = 2$  &$s = 3$  &$s = 4$  \\ \hline
  PSNR &31.96    &28.88    &27.35    \\ \hline
  SSIM &0.8972   &0.7993   &0.7276   \\ \hline
\end{tabular}
\label{table8}
\end{table}

\begin{figure*}
\centering
\subfigure{\includegraphics[width=1\textwidth]{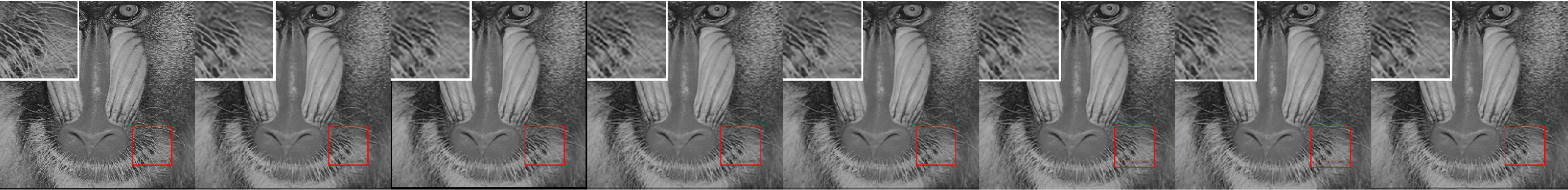} }
\subfigure{\includegraphics[width=1\textwidth]{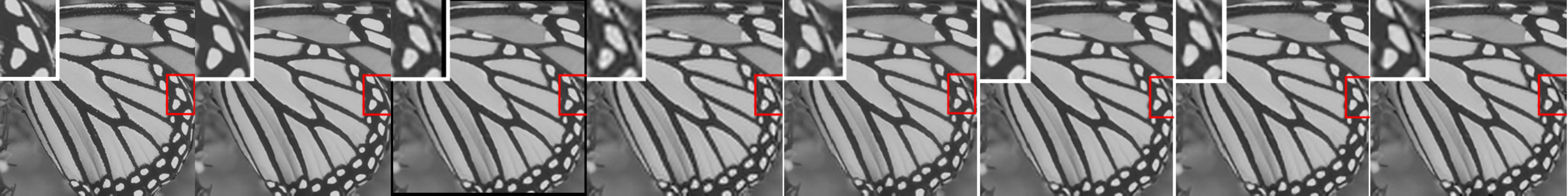} }
\subfigure{\includegraphics[width=1\textwidth]{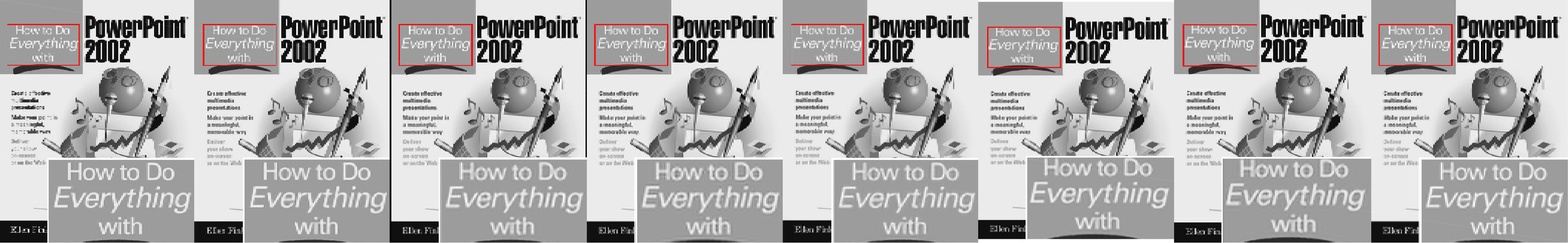} }
\subfigure{\includegraphics[width=1\textwidth]{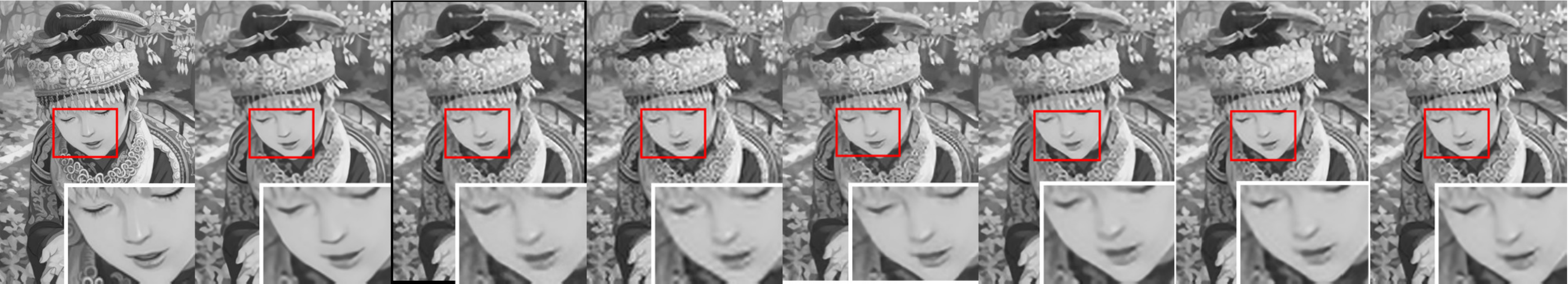} }
\caption{Visual results of image super-resolution.
Images from left to right column are: High resolution image;
the recovered image of RED30, ARFL+, CSC, CSCN, NBSRF, SRCNN, TSE.}
\label{fig14}
\end{figure*}

{\bf{Visual results}} Some visual results in grey-scale images are shown in Figure \ref{fig14}.
Note that it is straightforward to perform super-resolution on color images.

We can observe from
the second and third rows that our network is better at obtaining high resolution edges
and text. Meanwhile, our results seem much more smooth than others. For faces such as
the fourth row, out network still obtains better visually results.

\subsection{JPEG deblocking}

Lossy compression, such as JPEG, introduces complex compression artifacts,
particularly the blocking artifacts, ringing effects and blurring. In this
section, we carry out deblocking experiments to recover high quality images
from their JPEG compression. As in other compression artifacts reduction methods,
standard JPEG compression schemes of JPEG quality settings $q = 10$ and $q = 20$ in
MATLAB JPEG encoder are used. The LIVE1 dataset is used for evaluation, and we
have compared our method with AR-CNN \cite{DBLP:conf/iccv/DongDLT15}, SA-DCT
~\cite{DBLP:journals/tip/FoiKE07} and deeper SRCNN \cite{DBLP:conf/iccv/DongDLT15}.

The results are shown in Table \ref{table9}. We can observe that since the
Euclidean loss favors a high PSNR, our network outperforms other methods.
Compared to AR-CNN, the 30-layer network exceeds it by 0.37dB and 0.44dB on
compression quality of 10 and 20. Meanwhile, we can see that compared to
shallow networks, using significantly deeper networks does improve the deblocking performance.

\begin{table*}[htb!]
\centering
\caption{JPEG compression deblock: average PSNR results of LIVE1.}
\begin{tabular}{c|c c c c c c} \hline
              &SA-DCT   &Deeper SRCNN   &AR-CNN   &RED10   &RED20  &RED30           \\ \hline
  Quality $=10$  &28.65  	&28.92          &28.98    &29.24   &29.33  &\textbf{29.35}  \\ \hline
  Quality $=20$  &30.81  	&-              &31.29    &31.63   &31.71  &\textbf{31.73}  \\ \hline
\end{tabular}
\label{table9}
\end{table*}

\subsection{Non-blind deblurring}

\begin{figure*}
\centering
\subfigure{\includegraphics[width=1\textwidth]{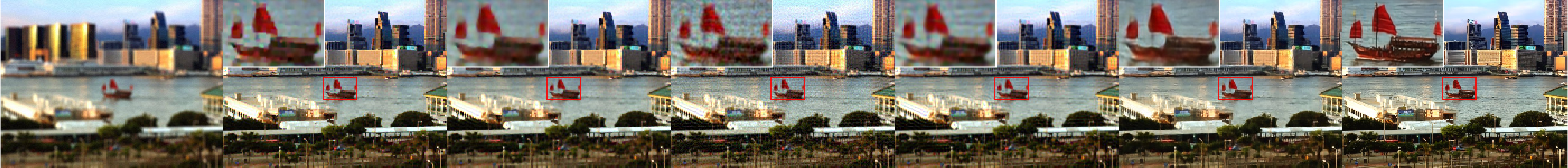} }
\subfigure{\includegraphics[width=1\textwidth]{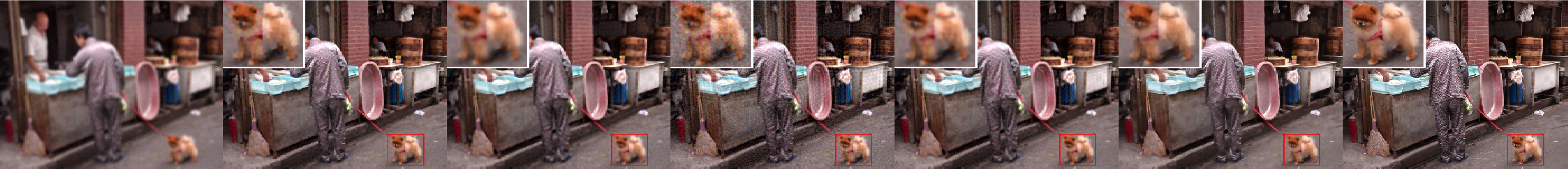} }
\subfigure{\includegraphics[width=1\textwidth]{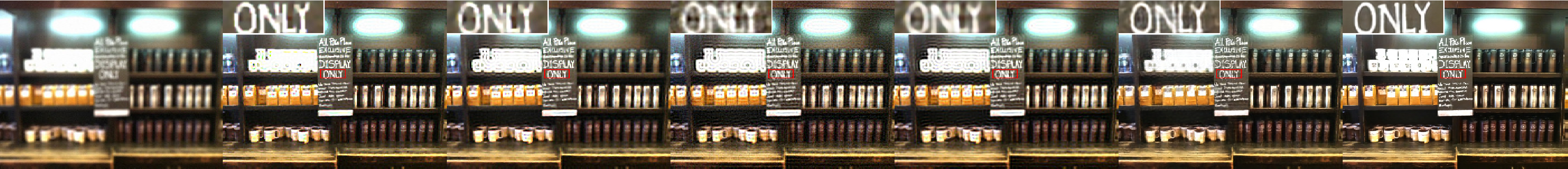} }
\caption{Visual comparisons on non-blind deblurring. Images from left to right are:
blurred images, the results of Cho~\cite{DBLP:conf/iccv/ChoWL11},
Krishnan~\cite{DBLP:conf/nips/KrishnanF09}, Levin~\cite{DBLP:journals/tog/LevinFDF07},
Schuler~\cite{DBLP:conf/cvpr/SchulerBHS13}, Xu~\cite{DBLP:conf/nips/XuRLJ14}
and our method.}
\label{fig17}
\end{figure*}

We mainly follow the experimental protocols as in ~\cite{DBLP:conf/nips/XuRLJ14} for
evaluation of non-blind deblurring. The performance on deblurring ``disk", ``motion"
and ``gaussian" kernels are compared, as shown in Table \ref{table10}. We generate
blurred image patches with the corresponding kernels, and train end-to-end
mapping with pairs of blurred and non-blurred image patches. As we can see from the results,
our network outperforms those compared methods with significant improvements. Figure~\ref{fig17} shows some visual comparisons.
We can observe from the visual examples
that our network works better than the compared methods on recovering the image details,
as well as achieving visually more appealing results on low frequency image contents.

\begin{table*}[htb]
\centering
\caption{PSNR results on non-blind deblurring.}
\begin{tabular}{c|c c c c c c} \hline
  kernel tpye  &Krishnan~et al.~\cite{DBLP:conf/nips/KrishnanF09}
  &Levin~et al.~\cite{DBLP:journals/tog/LevinFDF07}
               &Cho~et al.~\cite{DBLP:conf/iccv/ChoWL11}
               &Schuler et al.~\cite{DBLP:conf/cvpr/SchulerBHS13}
               &Xu~et al.~\cite{DBLP:conf/nips/XuRLJ14}    &RED30	\\ \hline
  disk    	&25.94  &24.54  &23.97  &24.67  &26.01  &\textbf{32.13}	\\ \hline
  motion    &30.34  &37.80  &33.25  &-      &-      &\textbf{38.84}	\\ \hline
  gaussian  &27.90  &32.34  &30.09  &30.97  &-      &\textbf{34.49}	\\ \hline
 \end{tabular}
\label{table10}
\end{table*}

\subsection{Image inpainting}
In this section,
we conduct text removal for experiments of image inpainting. Text is added
to the original image from the LIVE1 dataset  with font size of 10 and 20.
We have compared our method with FoE ~\cite{DBLP:journals/ijcv/RothB09}. For our model,
we extract image patches with text on them and learn a mapping from them to the
original patches. For FoE, we provide both images with text and masks indicating
which pixel is corrupted.

The average PSNR and SSIM for font size 10 and 20 on LIVE are:
38.24dB, 0.9869 and 34.99dB, 0.9828 using 30-layer RED-Net, and they are much
better than those of FoE, which are 34.59dB, 0.9762 and 31.10dB, 0.9510. For scratch
removal, we randomly draw scratch on the clean image and test with our network and FoE. The PSNR and SSIM for our network are 39.41dB and 0.9923, which is much better than 32.92dB and 0.9686 of FoE.

\begin{figure*}
\centering
\subfigure{\includegraphics[width=0.9\textwidth]{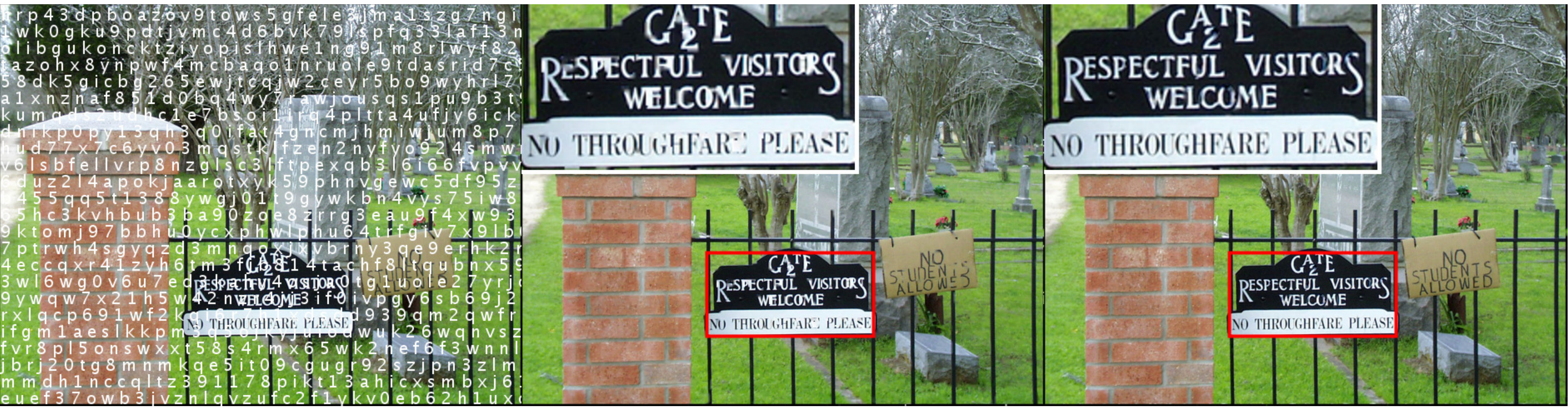} }
\subfigure{\includegraphics[width=0.9\textwidth]{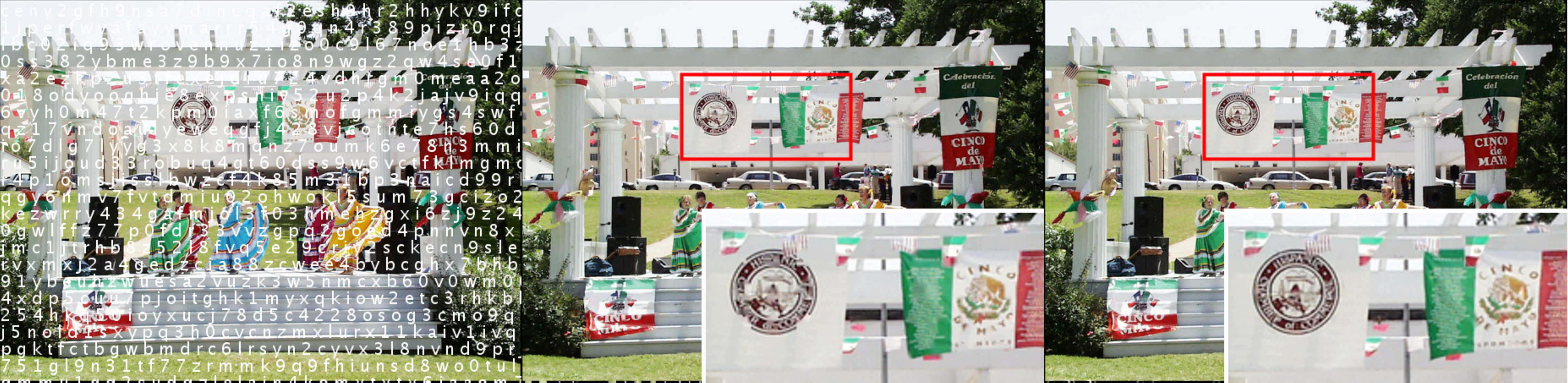} }
\subfigure{\includegraphics[width=0.9\textwidth]{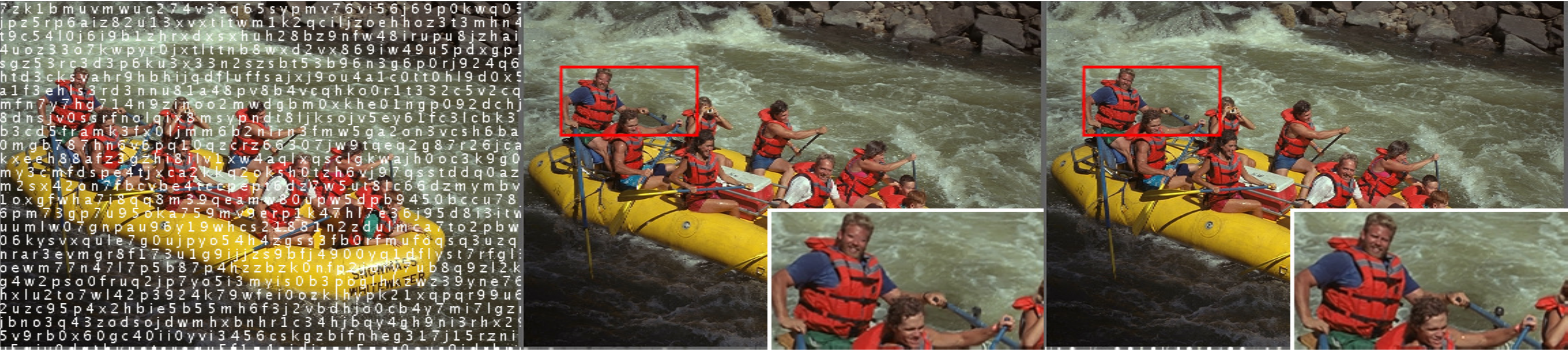} }
\subfigure{\includegraphics[width=0.9\textwidth]{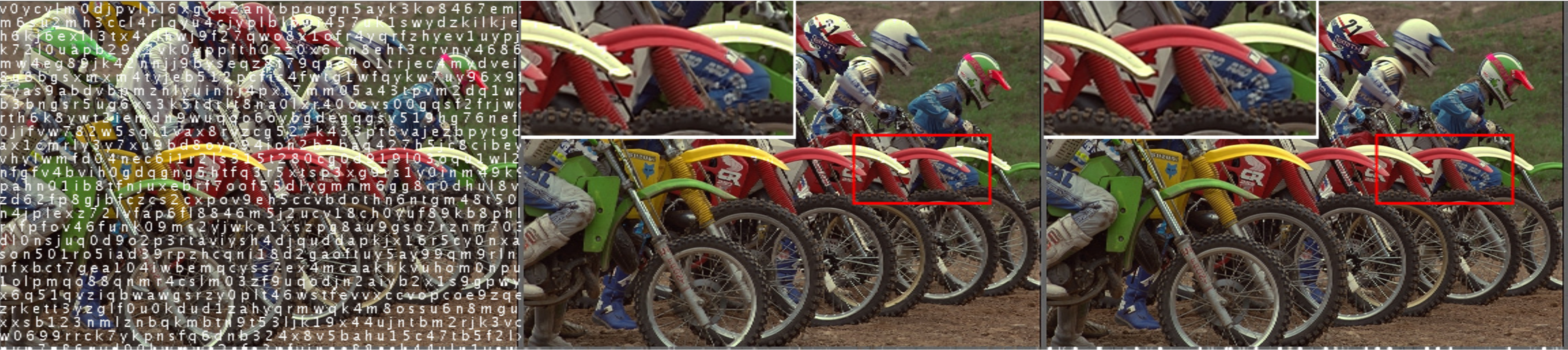} }
\subfigure{\includegraphics[width=0.9\textwidth]{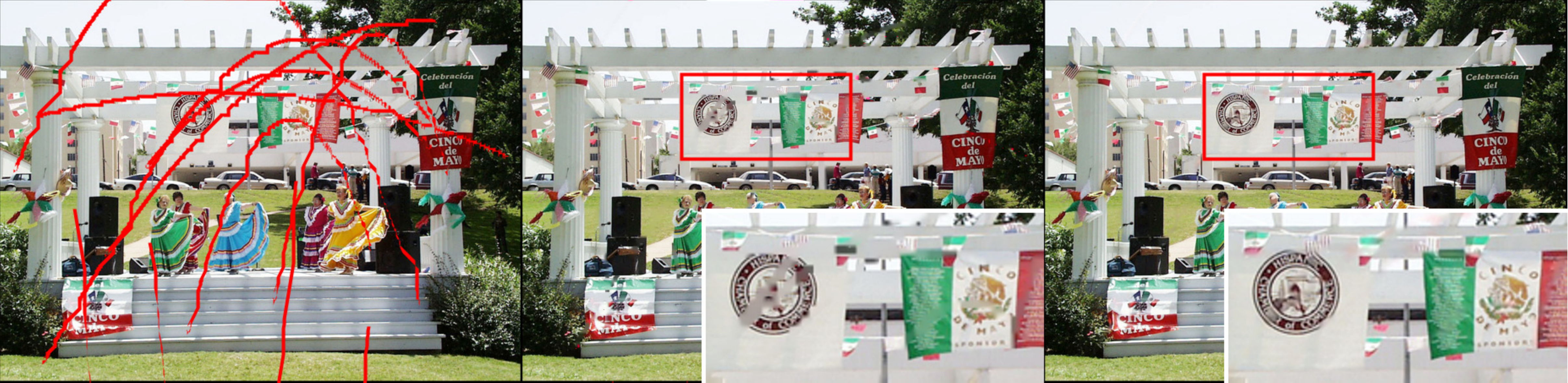} }
\caption{Visual results of our method and FoE. Images from left to right are:
Corrupted images, the inpainting results of FoE and the inpainting results of our method. We see better recovered details as shown in the zoomed
patches.}
\label{fig15}
\end{figure*}

Figure \ref{fig15} shows some visual comparisons of our method between FoE. We can
observe from the examples that our network is better at recovering text, logos,
faces and edges in the natural images. Looking on the first example, one may wonder
why the text in the original image is not eliminated. For traditional methods
such as FoE, this problem is addressed by providing a mask, which indicates the
location of corrupted pixels. While our network is trained on specific distributions
of corruptions, i.e., the text of font sizes 10 and 20 that are added.
It is equivalent to distinguishing corrupted and non-corrupted pixels of different distributions.

\section{Conclusions}

In this paper we have proposed a deep encoding and decoding framework
for image restoration. Convolution and deconvolution are combined,
modeling the restoration problem by extracting primary image content and recovering details.

More importantly,we propose to use skip connections, which helps on recovering
clean images and tackles the optimization difficulty caused by gradient
vanishing, and thus obtains performance gains when the network goes deeper.
Experimental results and our analysis show that our network achieves better
performance than state-of-the-art methods on image denoising,
image super-resolution, JPEG deblocking and image inpainting.

{
    \medskip
    \bibliographystyle{IEEEtran}
    \bibliography{CSRef}

\begin{thebibliography}{10}
\providecommand{\url}[1]{#1}
\csname url@samestyle\endcsname
\providecommand{\newblock}{\relax}
\providecommand{\bibinfo}[2]{#2}
\providecommand{\BIBentrySTDinterwordspacing}{\spaceskip=0pt\relax}
\providecommand{\BIBentryALTinterwordstretchfactor}{4}
\providecommand{\BIBentryALTinterwordspacing}{\spaceskip=\fontdimen2\font plus
\BIBentryALTinterwordstretchfactor\fontdimen3\font minus
  \fontdimen4\font\relax}
\providecommand{\BIBforeignlanguage}[2]{{%
\expandafter\ifx\csname l@#1\endcsname\relax
\typeout{** WARNING: IEEEtran.bst: No hyphenation pattern has been}%
\typeout{** loaded for the language `#1'. Using the pattern for}%
\typeout{** the default language instead.}%
\else
\language=\csname l@#1\endcsname
\fi
#2}}
\providecommand{\BIBdecl}{\relax}
\BIBdecl

\bibitem{DBLP:journals/corr/HeZRS15}
K.~He, X.~Zhang, S.~Ren, and J.~Sun, ``Deep residual learning for image
  recognition,'' in \emph{Proc. IEEE Conf. Comp. Vis. Patt. Recogn.}, 2016.

\bibitem{DBLP:conf/iccv/MairalBPSZ09}
J.~Mairal, F.~R. Bach, J.~Ponce, G.~Sapiro, and A.~Zisserman, ``Non-local
  sparse models for image restoration,'' in \emph{Proc. IEEE Int. Conf. Comp.
  Vis.}, 2009, pp. 2272--2279.

\bibitem{DBLP:journals/tip/DongZSL13}
W.~Dong, L.~Zhang, G.~Shi, and X.~Li, ``Nonlocally centralized sparse
  representation for image restoration,'' \emph{{IEEE} Trans. Image Process.},
  vol.~22, no.~4, pp. 1620--1630, 2013.

\bibitem{DBLP:journals/pami/SchmidtJNRR16}
U.~Schmidt, J.~Jancsary, S.~Nowozin, S.~Roth, and C.~Rother, ``Cascades of
  regression tree fields for image restoration,'' \emph{{IEEE} Trans. Pattern
  Anal. Mach. Intell.}, vol.~38, no.~4, pp. 677--689, 2016.

\bibitem{DBLP:conf/cvpr/SchmidtR14}
U.~Schmidt and S.~Roth, ``Shrinkage fields for effective image restoration,''
  in \emph{Proc. IEEE Conf. Comp. Vis. Patt. Recogn.}, 2014, pp. 2774--2781.

\bibitem{DBLP:conf/iccv/ZoranW11}
D.~Zoran and Y.~Weiss, ``From learning models of natural image patches to whole
  image restoration,'' in \emph{Proc. IEEE Int. Conf. Comp. Vis.}, 2011, pp.
  479--486.

\bibitem{DBLP:conf/cvpr/LiuXZG15}
H.~Liu, R.~Xiong, J.~Zhang, and W.~Gao, ``Image denoising via adaptive
  soft-thresholding based on non-local samples,'' in \emph{Proc. IEEE Conf.
  Comp. Vis. Patt. Recogn.}, 2015, pp. 484--492.

\bibitem{DBLP:conf/iccv/ChenZY15}
F.~Chen, L.~Zhang, and H.~Yu, ``External patch prior guided internal clustering
  for image denoising,'' in \emph{Proc. IEEE Int. Conf. Comp. Vis.}, 2015, pp.
  603--611.

\bibitem{DBLP:conf/iccv/XuZZZF15}
J.~Xu, L.~Zhang, W.~Zuo, D.~Zhang, and X.~Feng, ``Patch group based nonlocal
  self-similarity prior learning for image denoising,'' in \emph{Proc. IEEE
  Int. Conf. Comp. Vis.}, 2015, pp. 244--252.

\bibitem{DBLP:conf/cvpr/GuZZF14}
S.~Gu, L.~Zhang, W.~Zuo, and X.~Feng, ``Weighted nuclear norm minimization with
  application to image denoising,'' in \emph{Proc. IEEE Conf. Comp. Vis. Patt.
  Recogn.}, 2014, pp. 2862--2869.

\bibitem{DBLP:conf/accv/TimofteSG14}
R.~Timofte, V.~D. Smet, and L.~J.~V. Gool, ``{A+:} adjusted anchored
  neighborhood regression for fast super-resolution,'' in \emph{Proc. Asian
  Conf. Comp. Vis.}, 2014, pp. 111--126.

\bibitem{DBLP:conf/cvpr/YangLC13}
J.~Yang, Z.~Lin, and S.~Cohen, ``Fast image super-resolution based on in-place
  example regression,'' in \emph{Proc. IEEE Conf. Comp. Vis. Patt. Recogn.},
  2013, pp. 1059--1066.

\bibitem{DBLP:conf/cvpr/ZhuZY14}
Y.~Zhu, Y.~Zhang, and A.~L. Yuille, ``Single image super-resolution using
  deformable patches,'' in \emph{Proc. IEEE Conf. Comp. Vis. Patt. Recogn.},
  2014, pp. 2917--2924.

\bibitem{DBLP:conf/cvpr/ZhuZ0Y15}
Y.~Zhu, Y.~Zhang, B.~Bonev, and A.~L. Yuille, ``Modeling deformable gradient
  compositions for single-image super-resolution,'' in \emph{Proc. IEEE Conf.
  Comp. Vis. Patt. Recogn.}, 2015, pp. 5417--5425.

\bibitem{DBLP:conf/iccv/RieglerSRB15}
G.~Riegler, S.~Schulter, M.~R{\"{u}}ther, and H.~Bischof, ``Conditioned
  regression models for non-blind single image super-resolution,'' in
  \emph{Proc. IEEE Int. Conf. Comp. Vis.}, 2015, pp. 522--530.

\bibitem{DBLP:conf/iccv/GuZXMFZ15}
S.~Gu, W.~Zuo, Q.~Xie, D.~Meng, X.~Feng, and L.~Zhang, ``Convolutional sparse
  coding for image super-resolution,'' in \emph{Proc. IEEE Int. Conf. Comp.
  Vis.}, 2015, pp. 1823--1831.

\bibitem{DBLP:conf/iccv/WangLYHH15}
Z.~Wang, D.~Liu, J.~Yang, W.~Han, and T.~S. Huang, ``Deep networks for image
  super-resolution with sparse prior,'' in \emph{Proc. IEEE Int. Conf. Comp.
  Vis.}, 2015, pp. 370--378.

\bibitem{DBLP:conf/nips/XieXC12}
J.~Xie, L.~Xu, and E.~Chen, ``Image denoising and inpainting with deep neural
  networks,'' in \emph{Proc. Advances in Neural Inf. Process. Syst.}, 2012, pp.
  350--358.

\bibitem{DBLP:journals/ijcv/RothB09}
S.~Roth and M.~J. Black, ``Fields of experts,'' \emph{Int. J. Comput. Vision},
  vol.~82, no.~2, pp. 205--229, 2009.

\bibitem{DBLP:journals/tip/MairalES08}
J.~Mairal, M.~Elad, and G.~Sapiro, ``Sparse representation for color image
  restoration,'' \emph{{IEEE} Trans. Image Process.}, vol.~17, no.~1, pp.
  53--69, 2008.

\bibitem{DBLP:conf/iccv/DongDLT15}
C.~Dong, Y.~Deng, C.~C. Loy, and X.~Tang, ``Compression artifacts reduction by
  a deep convolutional network,'' in \emph{Proc. IEEE Int. Conf. Comp. Vis.},
  2015, pp. 576--584.

\bibitem{DBLP:journals/tip/FoiKE07}
A.~Foi, V.~Katkovnik, and K.~O. Egiazarian, ``Pointwise shape-adaptive {DCT}
  for high-quality denoising and deblocking of grayscale and color images,''
  \emph{{IEEE} Trans. Image Process.}, vol.~16, no.~5, pp. 1395--1411, 2007.

\bibitem{DBLP:conf/eccv/JancsaryNR12}
J.~Jancsary, S.~Nowozin, and C.~Rother, ``Loss-specific training of
  non-parametric image restoration models: {A} new state of the art,'' in
  \emph{Proc. Eur. Conf. Comp. Vis.}, 2012, pp. 112--125.

\bibitem{DBLP:conf/icml/VincentLBM08}
P.~Vincent, H.~Larochelle, Y.~Bengio, and P.~Manzagol, ``Extracting and
  composing robust features with denoising autoencoders,'' in \emph{Proc. Int.
  Conf. Mach. Learn.}, 2008, pp. 1096--1103.

\bibitem{DBLP:conf/nips/BengioLPL06}
Y.~Bengio, P.~Lamblin, D.~Popovici, and H.~Larochelle, ``Greedy layer-wise
  training of deep networks,'' in \emph{Proc. Advances in Neural Inf. Process.
  Syst.}, 2006, pp. 153--160.

\bibitem{DBLP:conf/nips/JainS08}
V.~Jain and H.~S. Seung, ``Natural image denoising with convolutional
  networks,'' in \emph{Proc. Advances in Neural Inf. Process. Syst.}, 2008, pp.
  769--776.

\bibitem{DBLP:conf/cvpr/LongSD15}
J.~Long, E.~Shelhamer, and T.~Darrell, ``Fully convolutional networks for
  semantic segmentation,'' in \emph{Proc. IEEE Conf. Comp. Vis. Patt. Recogn.},
  2015, pp. 3431--3440.

\bibitem{DBLP:journals/pami/DongLHT16}
C.~Dong, C.~C. Loy, K.~He, and X.~Tang, ``Image super-resolution using deep
  convolutional networks,'' \emph{{IEEE} Trans. Pattern Anal. Mach. Intell.},
  vol.~38, no.~2, pp. 295--307, 2016.

\bibitem{DBLP:journals/spm/Milanfar13}
P.~Milanfar, ``A tour of modern image filtering: New insights and methods, both
  practical and theoretical,'' \emph{{IEEE} Signal Process. Mag.}, vol.~30,
  no.~1, pp. 106--128, 2013.

\bibitem{DBLP:journals/tip/DabovFKE07}
K.~Dabov, A.~Foi, V.~Katkovnik, and K.~O. Egiazarian, ``Image denoising by
  sparse 3-d transform-domain collaborative filtering,'' \emph{{IEEE} Trans.
  Image Processing}, vol.~16, no.~8, pp. 2080--2095, 2007.

\bibitem{DBLP:journals/tip/WangYWCYH15}
Z.~Wang, Y.~Yang, Z.~Wang, S.~Chang, J.~Yang, and T.~S. Huang, ``Learning
  super-resolution jointly from external and internal examples,'' \emph{{IEEE}
  Trans. Image Process.}, vol.~24, no.~11, pp. 4359--4371, 2015.

\bibitem{DBLP:journals/tip/ChatterjeeM09}
P.~Chatterjee and P.~Milanfar, ``Clustering-based denoising with locally
  learned dictionaries,'' \emph{{IEEE} Trans. Image Process.}, vol.~18, no.~7,
  pp. 1438--1451, 2009.

\bibitem{Rudin:1992:NTV:142273.142312}
L.~I. Rudin, S.~Osher, and E.~Fatemi, ``Nonlinear total variation based noise
  removal algorithms,'' \emph{Phys. D}, vol.~60, no. 1-4, pp. 259--268,
  November 1992.

\bibitem{Chan05TV}
T.~Chan, S.~Esedoglu, F.~Park, and A.~Yip, ``Recent developments in total
  variation image restoration,'' in \emph{In Mathematical Models of Computer
  Vision}.\hskip 1em plus 0.5em minus 0.4em\relax Springer Verlag, 2005.

\bibitem{Oli09TV}
J.~Oliveira, J.~Bioucas-Dias, and M.~A.~T. Figueiredo, ``Adaptive total
  variation image deblurring: a majorization-minimization approach,''
  \emph{Signal Processing}, vol.~89, no.~9, pp. 2479--2493, September 2009.

\bibitem{DBLP:journals/tip/EladA06}
M.~Elad and M.~Aharon, ``Image denoising via sparse and redundant
  representations over learned dictionaries,'' \emph{{IEEE} Trans. Image
  Process.}, vol.~15, no.~12, pp. 3736--3745, 2006.

\bibitem{DBLP:journals/tip/DongZSW11}
W.~Dong, L.~Zhang, G.~Shi, and X.~Wu, ``Image deblurring and super-resolution
  by adaptive sparse domain selection and adaptive regularization,''
  \emph{{IEEE} Trans. Image Process.}, vol.~20, no.~7, pp. 1838--1857, 2011.

\bibitem{DBLP:journals/pami/KimK10}
K.~I. Kim and Y.~Kwon, ``Single-image super-resolution using sparse regression
  and natural image prior,'' \emph{{IEEE} Trans. Pattern Anal. Mach. Intell.},
  vol.~32, no.~6, pp. 1127--1133, 2010.

\bibitem{DBLP:journals/tip/YangWHM10}
J.~Yang, J.~Wright, T.~S. Huang, and Y.~Ma, ``Image super-resolution via sparse
  representation,'' \emph{{IEEE} Trans. Image Process.}, vol.~19, no.~11, pp.
  2861--2873, 2010.

\bibitem{DBLP:conf/eccv/CuiCSZC14}
Z.~Cui, H.~Chang, S.~Shan, B.~Zhong, and X.~Chen, ``Deep network cascade for
  image super-resolution,'' in \emph{Proc. Eur. Conf. Comp. Vis.}, 2014, pp.
  49--64.

\bibitem{DBLP:journals/tsp/SivakumarD93}
K.~Sivakumar and U.~B. Desai, ``Image restoration using a multilayer perceptron
  with a multilevel sigmoidal function,'' \emph{{IEEE} Trans. Signal
  Processing}, vol.~41, no.~5, pp. 2018--2022, 1993.

\bibitem{DBLP:conf/cvpr/BurgerSH12}
H.~C. Burger, C.~J. Schuler, and S.~Harmeling, ``Image denoising: Can plain
  neural networks compete with {BM3D}?'' in \emph{Proc. IEEE Conf. Comp. Vis.
  Patt. Recogn.}, 2012, pp. 2392--2399.

\bibitem{DBLP:conf/iccv/NohHH15}
H.~Noh, S.~Hong, and B.~Han, ``Learning deconvolution network for semantic
  segmentation,'' in \emph{Proc. IEEE Int. Conf. Comp. Vis.}, 2015, pp.
  1520--1528.

\bibitem{hong2015decoupled}
S.~Hong, H.~Noh, and B.~Han, ``Decoupled deep neural network for
  semi-supervised semantic segmentation,'' in \emph{Proc. Advances in Neural
  Inf. Process. Syst.}, 2015.

\bibitem{DBLP:journals/corr/SrivastavaGS15}
R.~K. Srivastava, K.~Greff, and J.~Schmidhuber, ``Training very deep
  networks,'' in \emph{Proc. Advances in Neural Inf. Process. Syst.}, 2015.

\bibitem{DBLP:conf/icml/NairH10}
V.~Nair and G.~E. Hinton, ``Rectified linear units improve restricted boltzmann
  machines,'' in \emph{Proc. Int. Conf. Mach. Learn.}, 2010, pp. 807--814.

\bibitem{jia2014caffe}
Y.~Jia, E.~Shelhamer, J.~Donahue, S.~Karayev, J.~Long, R.~Girshick,
  S.~Guadarrama, and T.~Darrell, ``{Caffe}: Convolutional architecture for fast
  feature embedding,'' in \emph{Proc.\ ACM Int.\ Conf. Multimedia}, 2014, pp.
  675--678.

\bibitem{DBLP:journals/corr/KingmaB14}
D.~P. Kingma and J.~Ba, ``Adam: {A} method for stochastic optimization,'' in
  \emph{Proc.\ Int.\ Conf.\ Learning Representations}, 2015.

\bibitem{MartinFTM01}
D.~Martin, C.~Fowlkes, D.~Tal, and J.~Malik, ``A database of human segmented
  natural images and its application to evaluating segmentation algorithms and
  measuring ecological statistics,'' in \emph{Proc. IEEE Int. Conf. Comp.
  Vis.}, vol.~2, July 2001, pp. 416--423.

\bibitem{DBLP:conf/eccv/ZeilerF14}
M.~D. Zeiler and R.~Fergus, ``Visualizing and understanding convolutional
  networks,'' in \emph{Proc. Eur. Conf. Comp. Vis.}, 2014.

\bibitem{DBLP:journals/corr/SermanetEZMFL13}
P.~Sermanet, D.~Eigen, X.~Zhang, M.~Mathieu, R.~Fergus, and Y.~LeCun,
  ``{OverFeat}: Integrated recognition, localization and detection using
  convolutional networks,'' \emph{Proc. Int. Conf. Learn. Representations},
  2014.

\bibitem{DBLP:journals/corr/SimonyanZ14a}
K.~Simonyan and A.~Zisserman, ``Very deep convolutional networks for
  large-scale image recognition,'' \emph{Proc. Int. Conf. Learn.
  Representations}, 2015.

\bibitem{DBLP:conf/iccv/SalvadorP15}
J.~Salvador and E.~Perez{-}Pellitero, ``Naive bayes super-resolution forest,''
  in \emph{Proc. IEEE Int. Conf. Comp. Vis.}, 2015, pp. 325--333.

\bibitem{DBLP:conf/cvpr/HuangSA15}
J.~Huang, A.~Singh, and N.~Ahuja, ``Single image super-resolution from
  transformed self-exemplars,'' in \emph{Proc. IEEE Conf. Comp. Vis. Patt.
  Recogn.}, 2015, pp. 5197--5206.

\bibitem{DBLP:conf/cvpr/SchulterLB15}
S.~Schulter, C.~Leistner, and H.~Bischof, ``Fast and accurate image upscaling
  with super-resolution forests,'' in \emph{Proc. IEEE Conf. Comp. Vis. Patt.
  Recogn.}, 2015, pp. 3791--3799.

\bibitem{DBLP:journals/corr/KimLL15b}
J.~Kim, J.~K. Lee, and K.~M. Lee, ``Accurate image super-resolution using very
  deep convolutional networks,'' in \emph{Proc. IEEE Conf. Comp. Vis. Patt.
  Recogn.}, 2016.

\bibitem{DBLP:journals/corr/KimLL15a}
------, ``Deeply-recursive convolutional network for image super-resolution,''
  in \emph{Proc. IEEE Conf. Comp. Vis. Patt. Recogn.}, 2016.

\bibitem{NIPS2016Mao}
X.~Mao, C.~Shen, and Y.~Yang, ``Image denoising using very deep fully
  convolutional encoder-decoder networks with symmetric skip connections,'' in
  \emph{Proc. Advances in Neural Inf. Process. Syst.}, 2016.

\bibitem{DBLP:conf/nips/XuRLJ14}
L.~Xu, J.~S.~J. Ren, C.~Liu, and J.~Jia, ``Deep convolutional neural network
  for image deconvolution,'' in \emph{Proc. Advances in Neural Inf. Process.
  Syst.}, 2014, pp. 1790--1798.

\bibitem{DBLP:conf/nips/KrishnanF09}
D.~Krishnan and R.~Fergus, ``Fast image deconvolution using hyper-laplacian
  priors,'' in \emph{Proc. Advances in Neural Inf. Process. Syst.}, 2009, pp.
  1033--1041.

\bibitem{DBLP:journals/tog/LevinFDF07}
A.~Levin, R.~Fergus, F.~Durand, and W.~T. Freeman, ``Image and depth from a
  conventional camera with a coded aperture,'' \emph{{ACM} Trans. Graph.},
  vol.~26, no.~3, p.~70, 2007.

\bibitem{DBLP:conf/iccv/ChoWL11}
S.~Cho, J.~Wang, and S.~Lee, ``Handling outliers in non-blind image
  deconvolution,'' in \emph{Proc. IEEE Int. Conf. Comp. Vis.}, 2011, pp.
  495--502.

\bibitem{DBLP:conf/cvpr/SchulerBHS13}
C.~J. Schuler, H.~C. Burger, S.~Harmeling, and B.~Sch{\"{o}}lkopf, ``A machine
  learning approach for non-blind image deconvolution,'' in \emph{Proc. IEEE
  Conf. Comp. Vis. Patt. Recogn.}, 2013, pp. 1067--1074.

\end{thebibliography}
}

\end{document}